\documentclass[10pt,journal,compsoc]{IEEEtran}
\usepackage[nocompress]{cite}

\usepackage[margin=0pt,font=small,labelfont=bf,labelsep=endash,tableposition=top]{caption}
\usepackage{epsfig}
\usepackage{graphicx}
\usepackage{amsmath}
\usepackage{amssymb}

\usepackage{multirow}
\usepackage{bm}
\usepackage{algorithm}
\usepackage{algorithmicx}
\usepackage{algpseudocode}
\usepackage{color}
\usepackage{enumitem}
\usepackage{arydshln}
\usepackage{pifont}
\usepackage{tablefootnote}
\usepackage{subfigure}
\usepackage{microtype}
\usepackage{url}
\usepackage{mathtools,xspace}
\usepackage{xcolor}
\usepackage{hyperref}

\def\etal{{\it et al.}\xspace}
\def\ie{{\it i.e.}\xspace}
\def\eg{{\it e.g.}\xspace}
\def\etc{{\it etc}\xspace}

\newlength\savedwidth
\newcommand\whline{\noalign{\global\savedwidth\arrayrulewidth
                           \global\arrayrulewidth 0.8pt}%
                  \hline
                  \noalign{\global\arrayrulewidth\savedwidth}}

\hypersetup{
    colorlinks=true,
    breaklinks=true,
    urlcolor= blue,
    linkcolor= red,
    bookmarksopen=false,
    filecolor=black,
    citecolor=blue,
    linkbordercolor=blue
}
\AtBeginDocument{\hypersetup{pdfborder={0 0 1}}}%

\begin{document}

\title{\textls[-13]{ABINet++: Autonomous, Bidirectional and Iterative Language Modeling for Scene Text Spotting}}

\author{Shancheng Fang,
        Zhendong Mao,
        Hongtao Xie,
        Yuxin Wang,
        Chenggang Yan,
        Yongdong Zhang
\vspace{-1em}
\IEEEcompsocitemizethanks{
\IEEEcompsocthanksitem S. Fang, Z. Mao, H. Xie, Y. Wang and Y. Zhang are with the School of Information Science and Technology, University of Science and Technology of China, Hefei, Anhui
230022, China. \protect 
E-mail:\{fangsc,zdmao,htxie,zhyd73\}@ustc.edu.cn, wangyx58@mail.ustc.edu.cn.
\vspace{-1em}
}

\IEEEcompsocitemizethanks{\IEEEcompsocthanksitem C. Yan is with the School of Automation, Hangzhou Dianzi University, Hangzhou 310018, China. \protect E-mail: cgyan@hdu.edu.cn
}
\thanks{(Corresponding authors: Zhendong Mao and Hongtao Xie.)}}
\markboth{Journal of \LaTeX\ Class Files}%
{Shell \MakeLowercase{\textit{et al.}}: Bare Demo of IEEEtran.cls for Computer Society Journals}

\IEEEtitleabstractindextext{%
\begin{abstract}

Scene text spotting is of great importance to the computer vision community due to its wide variety of applications. Recent methods attempt to introduce linguistic knowledge for challenging recognition rather than pure visual classification. However, how to effectively model the linguistic rules in end-to-end deep networks remains a research challenge. In this paper, we argue that the limited capacity of language models comes from 1) implicit language modeling; 2) unidirectional feature representation; and 3) language model with noise input. Correspondingly, we propose an autonomous, bidirectional and iterative ABINet++ for scene text spotting. Firstly, the autonomous suggests enforcing explicitly language modeling by decoupling the recognizer into vision model and language model and blocking gradient flow between both models. Secondly, a novel bidirectional cloze network (BCN) as the language model is proposed based on bidirectional feature representation. Thirdly, we propose an execution manner of iterative correction for the language model which can effectively alleviate the impact of noise input. Additionally, based on an ensemble of the iterative predictions, a self-training method is developed which can learn from unlabeled images effectively. Finally, to polish ABINet++ in long text recognition, we propose to aggregate horizontal features by embedding Transformer units inside a U-Net, and design a position and content attention module which integrates character order and content to attend to character features precisely. ABINet++ achieves state-of-the-art performance on both scene text recognition and scene text spotting benchmarks, which consistently demonstrates the superiority of our method in various environments especially on low-quality images. Besides, extensive experiments including in English and Chinese also prove that, a text spotter that incorporates our language modeling method can significantly improve its performance both in accuracy and speed compared with commonly used attention-based recognizers. Code is available at \url{https://github.com/FangShancheng/ABINet-PP}.

\end{abstract}

}

\maketitle

\IEEEdisplaynontitleabstractindextext

\IEEEpeerreviewmaketitle

\IEEEraisesectionheading{\section{Introduction}\label{sec:introduction}}

\IEEEPARstart{P}{ossessing} the capability of spotting text from scene images is indispensable to many applications of artificial intelligence, such as image retrieval, office automation, intelligent transportation system, \etc~\cite{ye2014text,long2020scene,wan2020vocabulary}. To this end, previous approaches either isolate the procedure of scene text spotting into text detection and text recognition~\cite{jaderberg2016reading}, or develop an end-to-end trainable network with the detection and recognition modules embedded~\cite{wang2021towards,liu2020abcnetv2}. Either way, the recognition of text plays a pivotal role as it transcribes pixels or features to character sequence fundamentally.

To recognize text from images, early attempts regard characters as meaningless symbols and read the symbols by classification models~\cite{wang2011end, jaderberg2016reading, lyu2018mask}. However, when confronted with challenging environments such as occlusion, blur, noise, \etc., it becomes faint due to out of visual discrimination. Fortunately, as text carries rich linguistic information, characters can be reasoned according to the context. Therefore, a bunch of methods~\cite{jaderberg2014deep,jaderberg2015deep,qiao2020seed, liao2020mask} turn their attention to language modeling and achieve undoubted improvement.

However, how to effectively model linguistic behavior in human reading is still an open problem. From the observations of psychology, we can make three assumptions about human reading that language modeling is autonomous, bidirectional and iterative: 1) as both deaf-mute and blind people could have fully functional vision and language separately, we use the term \emph{autonomous} to interpret the independence of learning between vision and language. The \emph{autonomous} also implies a good interaction between vision and language that independently learned language knowledge could contribute to the recognition of characters in vision. 2) The action of reasoning character context behaves like a cloze task since illegible characters can be viewed as blanks. Thus, prediction can be made using the cues of legible characters on the left side and right side of the illegible characters simultaneously, which is corresponding to the \emph{bidirectional}. 3) The \emph{iterative} describes that under challenging environments, humans adopt a progressive strategy to improve prediction confidence by iteratively correcting the recognized results.

\begin{figure}
   \begin{center}
      \includegraphics[width=0.48\textwidth]{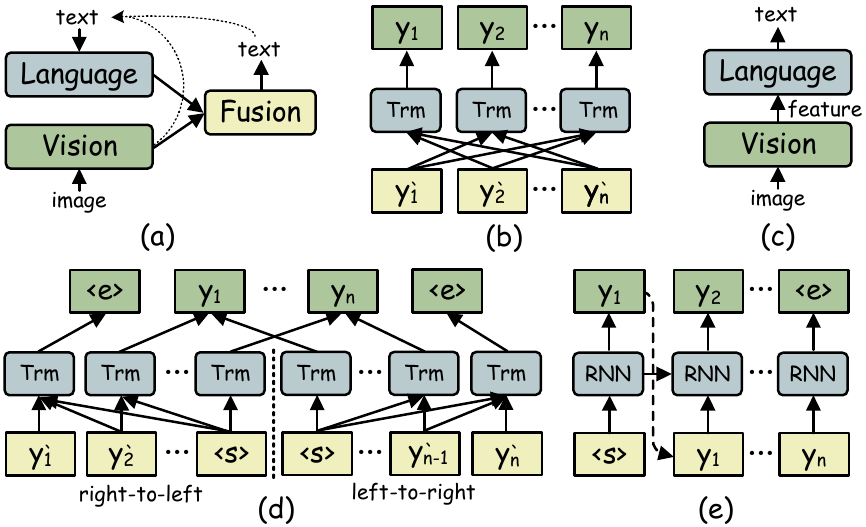}
      \caption{(a) Our autonomous language model with iterative correction. (b) Our bidirectional structure. (c) Coupled language model. (d) Ensemble of two unidirectional Transformers in parallel-prediction. (e) Unidirectional RNN in auto-regression. }
      \label{fig:overall}
   \end{center}
   \vspace{-2em}
\end{figure}

Firstly, applying the \textbf{autonomous} principle to scene text recognition means that recognition models should be decoupled into vision model~(VM) and language model~(LM), and the sub-models could be served as functional units independently and learned separately. Recent attention-based methods typically design LMs based on RNNs or Transformer~\cite{vaswani2017attention}, where the linguistic rules are learned \emph{implicitly} within a coupled model~\cite{lee2016recursive, shi2018aster, sheng2019nrtr} (Fig.~\ref{fig:overall}c). Nevertheless, whether and how well the LMs learn character relationship is unknowable. Besides, this kind of methods is infeasible to capture rich prior knowledge by directly pre-training LM from large-scale unlabeled text.

Secondly, compared with unidirectional LMs~\cite{sundermeyer2012lstm}, LMs with the \textbf{bidirectional} principle capture twice the amount of information. A straightforward way to construct a bidirectional model is to merge a left-to-right model and a right-to-left model~\cite{peters2018deep, devlin2018bert}, either at probability-level~\cite{wang2020decoupled,shi2018aster} or at feature-level~\cite{yu2020towards} (Fig.~\ref{fig:overall}d). However, they are strictly less powerful as their language features are unidirectional \emph{representation} in fact. Also, the ensemble models mean twice as expensive both in computations and parameters. A recent striking work in NLP is BERT~\cite{devlin2018bert}, which introduces a deep bidirectional representation learned by masking text tokens. Directly applying BERT to text recognition requires masking all the characters within a text instance, whereas this is extremely expensive since each time only one character can be masked.

Thirdly, LMs executed with the \textbf{iterative} principle can refine predictions obtained from visual and linguistic cues, which is not explored in current methods. The canonical way to perform an LM is auto-regression~\cite{wang2020decoupled,cheng2017focusing,wojna2017attention} (Fig.~\ref{fig:overall}e), in which error predictions are accumulated as noise and will be taken as the input for the following recognition. To adapt Transformer architectures, ~\cite{lyu20192d,yu2020towards} give up the auto-regression and adopt parallel-prediction (Fig.~\ref{fig:overall}d) to improve efficiency. However, the noise input still exists in the parallel-prediction where errors from VM output directly harm the accuracy of LM. In addition, the parallel-prediction in SRN~\cite{yu2020towards} suffers from an unaligned-length problem that SRN is tough to infer correct characters if text length is wrongly predicted by VM.

Considering the deficiencies of current methods from the aspects of internal interaction, feature representation and execution manner, we propose ABINet++ guided by the principles of \emph{Autonomous}, \emph{Bidirectional} and \emph{Iterative}. Firstly, we explore a decoupled method~(Fig.~\ref{fig:overall}a), where both VM and LM are autonomous units and could be pre-trained from images and text separately. Besides, the LM is enforced to learn linguistic rules explicitly by blocking gradient flow (BGF) between the VM and LM. Secondly, we design a novel bidirectional cloze network (BCN) as the LM~(Fig.~\ref{fig:overall}b), which eliminates the dilemma of combining two unidirectional models. The BCN is jointly conditioned on both left and right contexts, by specifying attention mask to control accessing of both side characters. Also, accessing across steps is not allowed to prevent leaking information. Thirdly, we propose an execution manner of iterative correction for the LM~(Fig.~\ref{fig:overall}a). By feeding the outputs of ABINet++ into the LM repeatedly, predictions can be refined progressively and the unaligned-length problem could be alleviated to a certain extent. Further, by treating the iterative predictions as an ensemble, a semi-supervised method is explored based on self-training, which exploits a new solution toward human-level recognition. Finally, for the VM in scene text recognition, we develop a position attention module to make visual predictions parallelly based on character order, which employs a U-Net to enhance feature representation for computing attention maps.

Based on the design philosophy of ABINet in our conference version~\cite{fang2021read}, we further endow it with the ability of end-to-end scene text spotting. Specifically, the major extensions include 1) without reconstructing the overall framework, we upgrade the VM by integrating Bezier curve detection~\cite{liu2020abcnet} to locate text in arbitrary shapes. The VM for text spotting inherits the attention mechanism from text the recognition model for character sequence prediction. 2) To improve the performance on long text images, we propose two approaches: horizontal feature aggregation and position \& content attention. The former argues aggregating information along the direction of character arrangement is essential for long text recognition, and thus we propose to use a U-Net with Transformer units embedded after the finest layer to aggregate features. Based on the position attention module, the latter combines it with a character content vector to exclude multiple response centers of attention maps, where the content vector is obtained by character prediction with an iterative strategy. 3) To enable efficient language learning within an end-to-end spotter, a text augmentation method for the LM is proposed which additionally samples external text and randomly misspells the sampled text.

Contributions of this paper mainly include: 
\begin{itemize}
\item
we propose autonomous, bidirectional and iterative principles to guide the design of LM for scene text recognition and spotting. Under these principles, the LM is a functional unit, which is required to extract bidirectional representation and correct prediction iteratively. 

\item
A novel BCN is introduced, which is able to learn the bidirectional representation of character sequence like cloze tasks, and thus can estimate the probability distribution of characters only using a single model. 

\item 
To improve the performance of the attention module in long text recognition, horizontal feature aggregation and position \& content attention are proposed. In the former we discuss that it is an essential factor to aggregate features along the direction of characters, and in the latter character content is introduced for learning better attention maps with accurate response centers.

\item 
To reinforce the learning of LM, we propose to use external text datasets (\eg, WikiText) for pre-training the LM, and sample misspelled text for training the LM online inside an end-to-end text spotter.

\item
The proposed ABINet++ achieves state-of-the-art performance on various text recognition benchmarks, and the ABINet++ trained with ensemble self-training shows promising improvement in realizing human-level recognition. Besides, extensive experiments on end-to-end text spotting benchmarks also demonstrate that the language modeling in ABINet++ can significantly improve the performance of mainstream text spotters, and thus achieve state-of-the-art performance.
\end{itemize}

\section{Related Work}
\label{sec:related-works}

\subsection{Scene Text Recognition}

Recent scene text recognition models fall into two groups: language-free and language-based methods. The former only considers visual discrimination, \eg text appearance including structure, shape, color, \etc. By contrast, the latter additionally learns linguistic rules. We note that currently the linguistic rules refer to spelling conventions as language-based methods mainly perform character-level context modeling, rather than capturing the semantic information of tokens/words~\cite{devlin2018bert, radford2018improving}.

\subsubsection{Language-free Methods} 

Language-free methods generally utilize visual features without the consideration of the relationship between characters, such as CTC-based~\cite{graves2006connectionist} and segmentation-based~\cite{li2017fully} methods. The CTC-based methods employ CNN to extract visual features and RNN to model features sequence. Then the CNN and RNN are trained end-to-end using CTC loss~\cite{shi2016end,he2016reading,su2017accurate,hu2020gtc}. The segmentation-based methods apply FCN to segment characters at pixel-level. Liao~\etal~\cite{liao2019mask} recognize characters by grouping the segmented pixels into text regions. Wan~\etal~\cite{wan2019textscanner} propose an additional order segmentation map which transcripts characters in correct order. Due to lacking linguistic information, the language-free methods cannot resolve the recognition problem in low-quality images commendably.

\subsubsection{Language-based Methods}

\noindent\textbf{Internal interaction between vision and language.} In some early works, the bags of $N$-grams of text string are predicted by a CNN which acts as an explicit LM~\cite{jaderberg2015deep,jaderberg2014deep,jaderberg2014synthetic}. After that the attention-based methods become popular, which implicitly model language rules using more powerful RNN~\cite{lee2016recursive,shi2018aster} or Transformer~\cite{wang2019simple,sheng2019nrtr}. The attention-based methods follow encoder-decoder architecture, where the encoder processes images and the decoder generates characters by focusing on relevant information from 1D image features~\cite{lee2016recursive,shi2016robust,shi2018aster,cheng2017focusing,cheng2018aon} or 2D image features~\cite{yang2017learning,wojna2017attention,liao2019scene, li2019show}. For example, R$^2$AM~\cite{lee2016recursive} employs recursive CNN as a feature extractor and LSTM as a learned LM implicitly modeling language at character-level, which avoids the use of $N$-grams. Further, this kind of methods is usually boosted by integrating a rectification module~\cite{shi2018aster,zhan2019esir,yang2019symmetry} for irregular images before feeding the images into networks. Different from the methods above, our method strives to build a more powerful LM by explicitly language modeling. In attempting to improve the language expression, some works introduce multiple losses where an additional loss comes from semantics~\cite{qiao2020seed, lyu20192d, yu2020towards, fang2018attention}. Among them, SEED~\cite{qiao2020seed} proposes to use a pre-trained FastText model to guide the training of RNN, which brings extra linguistic information. We deviate from this as our method directly pre-trains LM using unlabeled text, which is more feasible in practice.

\noindent\textbf{Representation of language features.} The character sequences in attention-based methods are generally modeled in left-to-right autoregressive way~\cite{lee2016recursive, shi2016robust, cheng2017focusing, yue2020robustscanner}. For instance, RobustScanner~\cite{yue2020robustscanner} inherits the unidirectional model of attention-based methods. Differently, they employ an additional position branch to enhance positional information and mitigate misrecognition in contextless scenarios. To utilize bidirectional information, methods like~\cite{graves2008novel, shi2018aster, wang2020decoupled, yu2020towards} use an ensemble model of two unidirectional models. Specifically, to capture global semantic context, SRN~\cite{yu2020towards} combines features from a left-to-right Transformer and a right-to-left Transformer for further prediction. We emphasize that the ensemble bidirectional model is intrinsically a unidirectional feature \emph{representation}.

Recent non-autoregressive methods in the NLP community are able to extract deep bidirectional features. For example, the masked language model (MLM) like BERT~\cite{devlin2018bert} is pre-trained by randomly replacing tokens with a special token {\tt{[MASK]}}. This masking strategy can also be applied to text recognition by replacing the mispredicted characters with {\tt{[MASK]}}~\cite{bhunia2021joint}. PIMNet~\cite{qiao2021pimnet} further extends it by gradually generating characters from a sequence of {\tt{[MASK]}} within multiple iterations. Another masking strategy is the attention-masking model (AMM). For example, by carefully designing the attention mask in self-attention, DisCo~\cite{kasai2020non} can predict tokens using a subset of the other reference tokens for machine translation, whereas an additional easy-first algorithm is needed for inference. Our BCN is also based on AMM. However, BCN proposes a diagonal mask operated in cross-attention, simultaneously keeping a concise structure and powerful performances both in effectiveness and efficiency.

\noindent\textbf{Execution manner of language models.} Currently, the network architectures of language models are mainly based on RNN and Transformer~\cite{vaswani2017attention}. The RNN-based LM is usually executed in auto-regression~\cite{wang2020decoupled,cheng2017focusing,wojna2017attention}, which takes the prediction of last character as input. Typical work such as DAN~\cite{wang2020decoupled} obtains the visual features of each character firstly using its proposed convolutional alignment module. After that GRU predicts each character by taking the prediction embedding of the last time step and the character feature of the current time step as input. The Transformer-based methods have superiority in parallel execution, where the inputs of each time step are either visual features~\cite{lyu20192d} or character embedding from the prediction of visual feature~\cite{yu2020towards}. Our method falls into the parallel execution, but we try to alleviate the issue of noise input existing in parallel LM.

\begin{figure*}
   \begin{center}
      \includegraphics[width=1.0\textwidth]{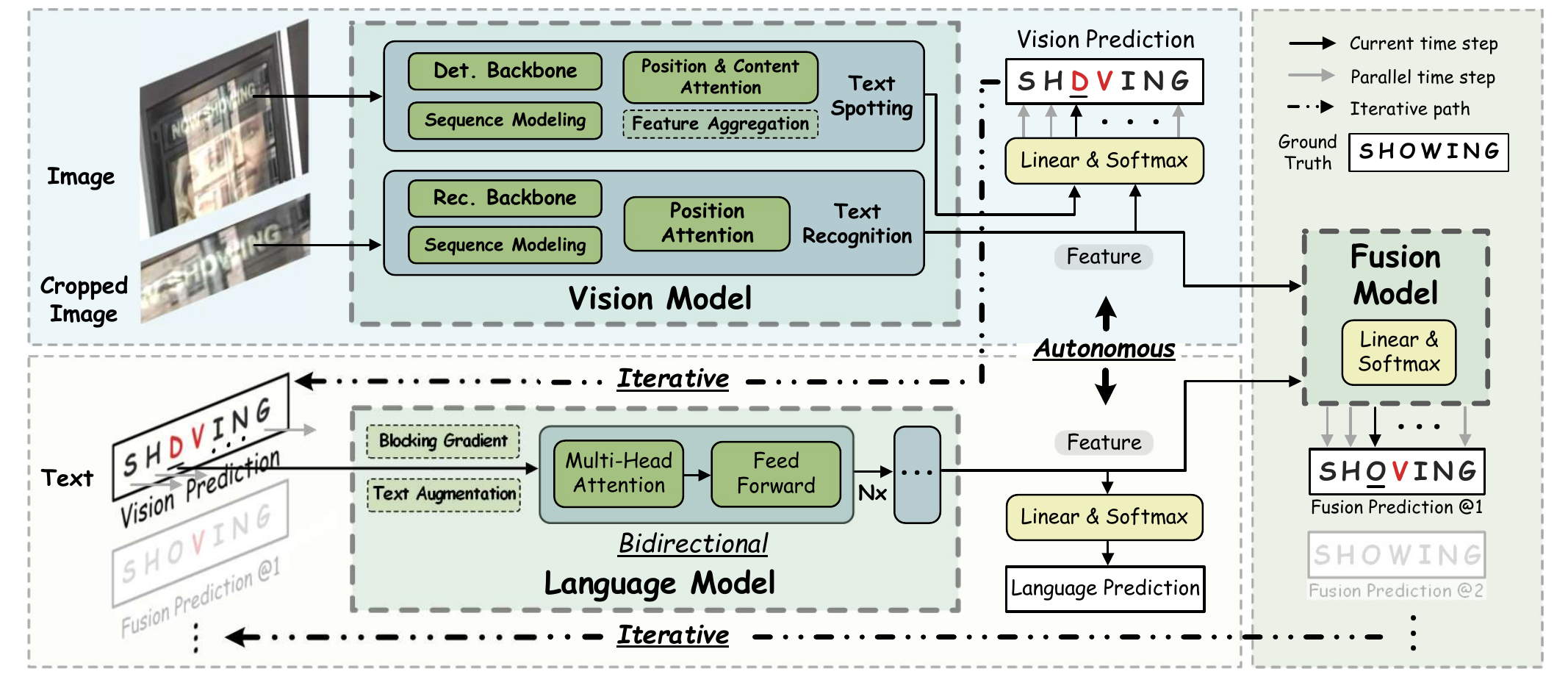}
      \caption{\textbf{A schematic overview of ABINet++.}}
      \label{fig:framework}
   \end{center}
   \vspace{-2em}
 \end{figure*}

\subsection{End-to-End Scene Text Spotting} 

Due to the serious challenges of scene text spotting, previous approaches try to tackle it by introducing two standalone models: text detection and text recognition. However, as inherent correlation exists in text detection and recognition, \eg shared feature learning and mutually performance boosting, recent methods explore a unified optimization scheme based on end-to-end trainable networks, either based on a two-stage framework that recognition is performed serially following detection, or a one-stage framework that simultaneously accomplishing detection and recognition. 

\subsubsection{Two-stage Text Spotter}

A critical point to bridging text detection and recognition models is to gather recognition features from shared features using detection predictions. To this end, various kinds of RoI extractors are employed, such as regularly-shaped extractors (\eg, RoI Pooling, ROIAlign) and arbitrarily-shaped extractors (\eg, Thin-Plate-Spline transformation, BezierAlign).

\noindent\textbf{Regularly-shaped Scene Text Spotter.} Li~\etal~\cite{li2017towards} make an attempt to jointly optimize the text detection and recognition networks. Based on the Faster R-CNN framework, this method additionally integrates a CTC-based recognition network serially after a rectangular detection network. The text features for detection and recognition are extracted by an RoI Pooling layer together with an LSTM module for varying-length feature modeling. With the aid of 2D attention recognition model, the enhanced version~\cite{wang2021towards} further improves its robustness on irregular text. Meanwhile, Busta~\etal~\cite{busta2017deep} propose a similar framework that detects rotated bounding boxes and extends RoI Pooling with bilinear sampling. Further, Liu~\etal~\cite{liu2018fots} propose FOTS aiming to settle the problem of multi-oriented recognition. FOTS introduces a new operation dubbed RoIRotate, which transforms oriented features to axis-aligned representation from quadrilateral detection results. Akin to RoIRotate, Text-Alignment is proposed by He~\etal~\cite{he2018end} to sample rotated features for text recognition. By incorporating an attention-based sequence-to-sequence recognizer, this method improves the performance largely compared with previous methods.

\noindent\textbf{Arbitrarily-shaped Scene Text Spotter.} To endow the extractor with the ability of arbitrarily-shaped representation, semantic segmentation based methods~\cite{lyu2018mask,qin2019towards,wang2021pan++} are investigated as their flexible pixel combination can form arbitrary descriptions. Lyu and Liao~\etal~\cite{lyu2018mask} propose Mask TextSpotter inspired by Mask R-CNN. The Mask TextSpotter utilizes RoIAlign as the feature extractor and further recognizes arbitrarily-shaped text using word segmentation and character segmentation. As Mask TextSpotter requires character-level supervision, the improved version~\cite{liao2019mask} eliminates this limitation by incorporating a spatial attention model. Further, Liao~\etal~\cite{liao2020mask} propose a segmentation proposal network in Mask TextSpotter v3 to replace RPN, enabling a hard binary-masking on RoI features. A similar operation called RoI masking is also adopted by Qin~\etal~\cite{qin2019towards}, where an attention-based recognizer takes the RoI features masked by a probability map as input. PAN++ proposed by Wang~\etal~\cite{wang2021pan++} first predicts text regions, text center kernels, and instance vectors simultaneously. Then complete text instances are recovered, and thereby the Masked RoI features for the text instances are fed into an attention-based recognizer. Different from the segmentation-based methods, another group of methods directly extends the extractor based on key-points regression~\cite{sun2018textnet,feng2019textdragon,wang2020all,qiao2020text,liu2020abcnet}. Sun~\etal~\cite{sun2018textnet} predict quadrilateral bounding boxes and rectify the quadrilateral features using perspective RoI transformation. Feng~\etal~\cite{feng2019textdragon} develop RoISlide to reconstruct a text line from a series of feature segments. Wang~\etal~\cite{wang2020all} and Qiao~\etal~\cite{qiao2020text} both regress fiducial points and rectify irregular text through Thin-Plate-Spline transformation. CRAFT~\cite{baek2020character} uses character region maps supervised by character-level annotations to help attention-based recognizer attend to precise character center points. Liu~\etal~\cite{liu2020abcnet} propose ABCNet representing a text instance by parameterized Bezier curves, whose control points are regressed by a single-shot and anchor-free detector. Then BezierAlign is developed to transform arbitrarily-shaped text features into horizontal representation, which can reduce the burden of recognition. ABCNet v2~\cite{liu2020abcnetv2} further strengthens the backbone using BiFPN and replaces the CTC-based recognizer with an attention-based recognizer, resulting in a significant improvement in accuracy. In our work, the BezierAlign is chosen as our RoI extractor due to its concise representation and seamless integration with our VM. The above text spotters mainly use CTC-based methods or attention-based RNN models as the recognizer. Differently, our work, to the best of our knowledge, is the first method to introduce an explicit LM for text spotting in the view of multi-modal fusion, and shows obvious advantages over previous recognizers.

\subsubsection{One-stage Text Spotter}

Recently, several works turn to a simple yet powerful one-stage structure, aiming at liberating text spotters from RoI extractors. Xing~\etal~\cite{xing2019convolutional} propose CharNet which can parallelly process detection and recognition in one pass. CharNet views the character as a basic unit whereas requires character-level annotations. Qiao~\etal~\cite{qiao2021mango} propose MANGO where text information is aggregated by an instance-level mask attention module and a character-level mask attention module. Nevertheless, this method needs a pre-training stage on datasets with character-level annotations. Wang~\etal~\cite{wang2021pgnet} propose PGNet which directly predicts four kinds of text segmentation maps. Further, PGNet gives up character-level annotations by developing a Point Gathering CTC loss. The one-stage methods remove the RoI extractor from the text spotter. However, their recognizers suffer from noise content due to inaccurate text localization.


\section{Scene Text Recognition}
\label{sec:str}

\subsection{Vision Model for Text Recognition}
\label{sec:vision}

\begin{figure}
   \begin{center}
      \includegraphics[width=0.5\textwidth]{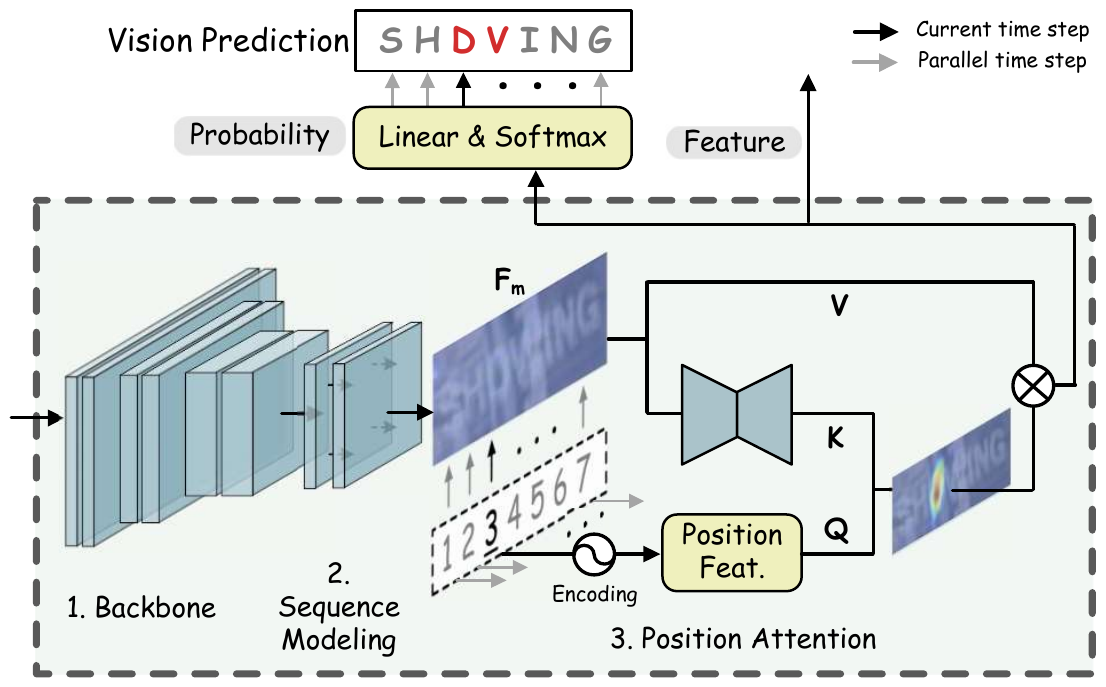}
      \caption{\textbf{Architecture of vision model for text recognition.}}
      \label{fig:vision}
   \end{center}
   \vspace{-0.5em}    
\end{figure}

As the procedure defined in~\cite{baek2019wrong}, the vision model (Fig.~\ref{fig:vision}) can be formulated as three stages: feature extraction, sequence modeling and character prediction. Following the previous methods, ResNet\footnote{There are 5 residual blocks in total and down-sampling is performed after the 1st and 3rd blocks.}~\cite{shi2018aster, wang2020decoupled} is employed as the backbone network $\mathcal{B}$ for feature extraction, and Transformer~\cite{yu2020towards, lyu20192d} is chosen as the sequence modeling network (SMN, denoted as $\mathcal{M}$). Besides, we propose a position attention (PA) module as the prediction network. Therefore, for image $\bm{x}$ we have:
\begin{align}
\mathbf{F}_b = \mathcal{B}(\bm{x}) \in \mathbb{R}^{h \times w \times d}, \label{eq:vision} \\
\mathbf{F}_m = \mathcal{M}(\mathbf{F}_b) \in \mathbb{R}^{h \times w \times d}, 
\end{align}
where $h,w$ are the size of the feature maps, which is equal to the 4x downsampling size of the image height and width. $d$ is the feature dimension.

The PA module transcribes visual features into character probabilities in parallel, which is based on the query paradigm~\cite{vaswani2017attention}:
\begin{align}
\mathbf{F}_v = \text{softmax}(\frac{\mathbf{Q}\mathbf{K}^\mathsf{T}}{\sqrt{d}})\mathbf{V}, \\
\mathbf{Q} = PE(t) \in \mathbb{R}^{t \times d}, \label{eq:query}\\
\mathbf{K} = \mathcal{G}(\mathbf{F}_m) \in \mathbb{R}^{h \times w \times d}, \\
\mathbf{V} = \mathcal{H}(\mathbf{F}_m) \in \mathbb{R}^{h \times w \times d}.
\end{align}
Concretely, $PE(\cdot)$ is the positional encodings~\cite{vaswani2017attention} of character order and $t$ is the length of character sequence. $\mathcal{G}(\cdot)$ is implemented by a U-Net like network~\cite{ronneberger2015u}, whose detailed structure is given in Table~\ref{tab:unet}. $\mathcal{H}(\cdot)$ is an identity mapping. 

\begin{table}[!t]
   \centering
   \newcommand{\tabincell}[2]{\begin{tabular}{@{}#1@{}}#2\end{tabular}}
   \caption{Structure of the U-Net, where both the encoder and decoder have 4 convolution layers with 3×3 kernels and 1 padding. The encoder performs downsampling by directly adjusting convolution strides, and the decoder achieves upsampling using interpolation layers. Skip connections are omitted for simplicity and features are combined using $add$ operation.}
   \label{tab:rec_struct}
   \footnotesize
   \resizebox{1.0\linewidth}{!}{
   \begin{tabular}{c|c|c|c|c}
     \whline
     \multirow{3}{*}{Layers} & \multicolumn{2}{c|}{Text Recognition} & \multicolumn{2}{c}{Text Spotting} \\
     \cline{2-5}
     & \tabincell{c}{Parameters \\ (Filters, Strides)} & \tabincell{c}{Output \\ ($H$, $W$)} & \tabincell{c}{Parameters \\ (Filters, Strides)} & \tabincell{c}{Output \\ ($H$, $W$)} \\
     \hline
     Conv. & $[64, (1,2)] \times 1$ & ($h$, $\frac{w}{2}$) & $[256, (2,2)] \times 1$ & ($\frac{h}{2}$, $\frac{w}{2}$) \\
     Conv. & $[64, (2,2)] \times 1$ & ($\frac{h}{2}$, $\frac{w}{4}$) & $[256, (2,2)] \times 1$ & ($\frac{h}{4}$, $\frac{w}{4}$) \\
     Conv. & $[64, (2,2)] \times 1$ & ($\frac{h}{4}$, $\frac{w}{8}$) & $[256, (2,2)] \times 1$ & ($\frac{h}{8}$, $\frac{w}{8}$) \\
     Conv. & $[64, (2,2)] \times 1$ & ($\frac{h}{8}$, $\frac{w}{16}$) & $[256, (1,1)] \times 1$ & ($\frac{h}{8}$, $\frac{w}{8}$) \\
     \hline
     Trans. & - & - & $\times 4$ & ($\frac{h}{8}$, $\frac{w}{8}$) \\
     \hline
     Up + Conv. & $[64, (1,1)] \times 1$ & ($\frac{h}{4}$, $\frac{w}{8}$) & $[256, (1,1)] \times 1$ & ($\frac{h}{8}$, $\frac{w}{8}$) \\
     Up + Conv. & $[64, (1,1)] \times 1$ & ($\frac{h}{2}$, $\frac{w}{4}$)  & $[256, (1,1)] \times 1$ & ($\frac{h}{4}$, $\frac{w}{4}$)\\
     Up + Conv. & $[64, (1,1)] \times 1$  & ($h$, $\frac{w}{2}$)  & $[256, (1,1)] \times 1$ & ($\frac{h}{2}$, $\frac{w}{2}$)\\
     Up + Conv. & $[512, (1,1)] \times 1$  & ($h$, $w$)  & $[256, (1,1)] \times 1$ & ($h$, $w$)\\
     \whline
   \end{tabular}}
   \label{tab:unet}
   \vspace{-1em}   
 \end{table}

\subsection{Language Model}
\subsubsection{Autonomous Strategy}
\label{sec:autonomous}

As shown in Fig.~\ref{fig:framework}, the autonomous strategy includes the following characteristics: 1) the LM is regarded as an independent model of spelling correction which takes probability vectors of characters as input and outputs probability distributions of expected characters. 2) The flow of the training gradient is blocked (BGF) at input vectors. 3) The LM could be trained separately from unlabeled text data.

Given a text string $\bm{y}_{1:n}=\{y_1, \ldots, y_n\}$ with length $n$, the attention-based methods~\cite{lee2016recursive,shi2018aster,cheng2017focusing,li2019show,sheng2019nrtr} basically perform language modeling implicitly, \ie, $P(y_i|\bm{y}_{1:{i-1}}, \mathcal{\bm{F}}_{v})$, where $\mathcal{F}_{v}$ is the visual features. The $\mathcal{F}_{v}$ might be a biased factor from the aspect of language modeling, as $P(y_i)$ is estimated partly conditioned on $\mathcal{F}_{v}$. Oppositely, the autonomous strategy defines an explicit LM which exactly achieves $P(y_i|\bm{y}_{1:{i-1}})$ by decoupling the VM and LM. On the other hand, as the context characters $\bm{y}_{1:{i-1}}$ are directly computed from $\mathcal{F}_{v}$ in our model, the biased factor still exists in backpropagation. Thus, we utilize BGF to isolate the learning between different modalities, which ensures the independence of vision and language learning.

Following the autonomous strategy, the ABINet++ can be divided into interpretable units. By taking the probability as input, LM could be replaceable (\ie, replaced with other language models directly, as in Table~\ref{tab:bidirectional}) and flexible (\eg, executed iteratively as in Section~\ref{sec:iterative}). Besides, the decoupled structure enforces LM to learn linguistic knowledge inevitably, which is radically distinguished from implicit language modeling where what the models exactly learn is unknowable. Furthermore, the autonomous strategy allows us to directly share the advanced progress in the NLP community. For instance, pre-training the LM can be an effective way to boost performance.

\subsubsection{Bidirectional Representation}

\begin{figure}
   \begin{center}
      \includegraphics[width=0.4\textwidth]{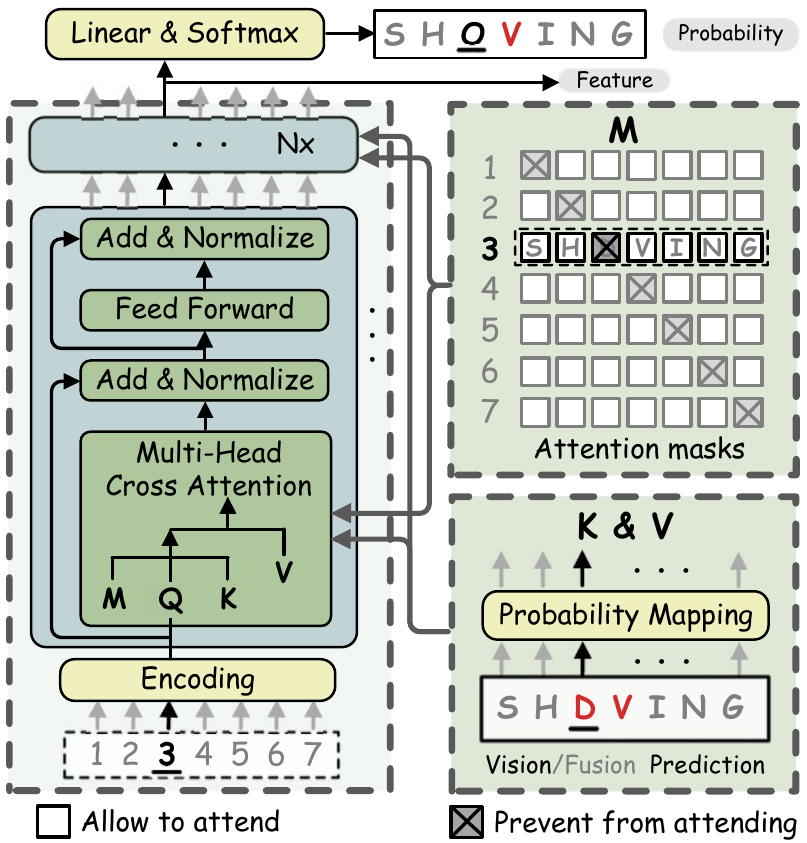}
      \caption{\textbf{Architecture of language model (BCN).}}
      \label{fig:label}
   \end{center}
   \vspace{-2em}   
\end{figure}

The conditional probability of $y_i$ for bidirectional and unidirectional models are $P(y_i|\bm{y}_{1:{i-1}},\bm{y}_{{i+1}:n})$ and $P(y_i|\bm{y}_{1:{i-1}})$, respectively. From the perspective of information theory, available entropy of a bidirectional representation can be quantified as $H_{\bm{y}} = (n-1)\log{c}$, where $c$ is the class number. However, for a unidirectional representation the information is $\frac{1}{n}\sum^n_{i=1}{(i-1)\log{c}}=\frac{1}{2}H_{\bm{y}}$. Our insight is that previous methods typically use an ensemble model of two unidirectional models, which essentially are unidirectional representations. The unidirectional representation basically captures $\frac{1}{2}H_{\bm{y}}$ information, resulting in limited capability in feature abstraction compared with the bidirectional counterpart.

Benefitting from the autonomous design in Section~\ref{sec:autonomous}, off-the-shelf NLP models with the ability of spelling correction can be transferred. A plausible way is to utilize the MLM strategy in BERT~\cite{devlin2018bert} by replacing $y_i$ with token {\tt{[MASK]}}. However, we notice that this is inefficient as the BERT should be separately called $n$ times for each text instance. Instead of masking the input characters, we propose BCN by specifying attention mask.

Overall, BCN is a variant of an $L$-layer Transformer decoder. Each layer of BCN is a series of multi-head attention (cross-attention) and feed-forward network~\cite{vaswani2017attention} followed by residual connection~\cite{he2016deep} and layer normalization~\cite{ba2016layer}, as shown in Fig.~\ref{fig:label}. Different from vanilla Transformer, character vectors are fed into the multi-head attention blocks rather than the first layer of the network. In addition, attention mask in multi-head attention is designed to prevent from “seeing itself". Besides, no self-attention is applied in BCN to avoid leaking information across time steps. The attention operation inside multi-head blocks can be formalized as:
\begin{align}
\mathbf{M}_{ij} &= \begin{cases} 0, & i \neq j \\ -\infty, & i = j \end{cases}, \label{eq:att:mask} \\
\mathbf{K}_i &= \mathbf{V}_i = P(y_i) \mathbf{W}_l,  \\
\mathbf{F}_{mha} &= \text{softmax}(\frac{\mathbf{Q}\mathbf{K}^\mathsf{T}}{\sqrt{d}} + \mathbf{M})\mathbf{V},
\end{align}
where $\mathbf{Q} \in \mathbb{R}^{t \times d}$ is the positional encodings of character order in the first layer and the outputs of the last layer otherwise. $\mathbf{K}, \mathbf{V} \in \mathbb{R}^{t \times d}$ are obtained from character probability $P(y_i) \in \mathbb{R}^{c}$, and $\mathbf{W}_l \in \mathbb{R}^{c \times d}$ is a learnable mapping matrix. $\mathbf{M} \in \mathbb{R}^{t \times t}$ is the attention mask which prevents from attending to the current character. After stacking BCN layers into deep architecture, the bidirectional representation $\mathbf{F}_{l}$ for the text $\bm{y}$ is determined.

By specifying the attention mask in a cloze fashion, BCN is able to learn more powerful bidirectional representation elegantly than the ensemble of unidirectional representation. Besides, benefitting from Transformer-like architecture, BCN can perform computation independently and parallelly. Also, it is more efficient than the ensemble models as only half of the computations and parameters are needed.

\begin{figure}
   \begin{center}
      \includegraphics[width=0.5\textwidth]{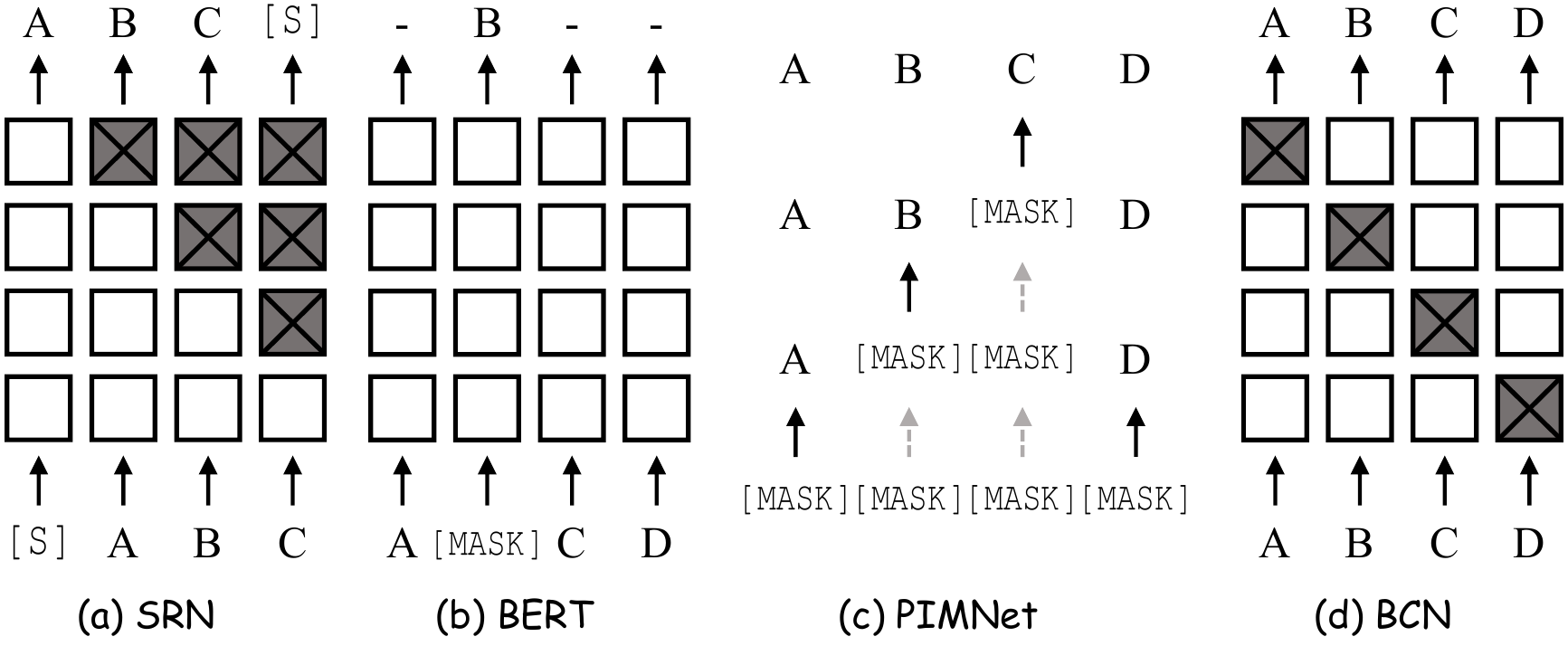}
      \caption{\textbf{Comparison of different language models.}}
      \label{fig:comparisons_lms}
   \end{center}
   \vspace{-1.5em}   
\end{figure}

Fig.~\ref{fig:comparisons_lms} gives a comparison of several LMs proposed recently. SRN is a unidirectional feature model, which adopts casual masks~\cite{radford2018improving} operated in self-attention. Differently, BERT, PIMNet and BCN can obtain bidirectional feature representation. Among them, the fundament of BERT and PIMNet is MLM, which generally needs multiple network passes for character decoding. By contrast, the characters in BCN are predicted in one shot, using a diagonal mask operated in cross-attention. A more detailed comparison can be found in the supplementary material.

\subsubsection{Iterative Correction}
\label{sec:iterative}

The parallel-prediction of Transformer takes noise inputs which typically are approximations from visual prediction~\cite{yu2020towards} or visual feature~\cite{lyu20192d}. Concretely, as the example shown in Fig.~\ref{fig:framework}, the desired condition of $P(\text{``O"})$ is ``SH-WING" in the bidirectional representation. However, due to the blurred and occluded environments, the actual condition obtained from VM is ``SH-VING", in which ``V" becomes noise and harms the confidence of prediction. It tends to be more hostile for LM with increasing error predictions in VM.

To cope with the problem of noise inputs, we propose iterative LM (illustrated in Fig.~\ref{fig:framework}). The LM is executed $M$ times repeatedly with different assignments for $\bm{y}$. For the first iteration, $\bm{y}_{i=1}$ is the probability prediction from the VM. For the subsequent iterations, $\bm{y}_{i \ge 2}$ is the probability prediction from the fusion model (Section~\ref{sec:fusion}) in last iteration. By this way, the LM can correct the vision prediction iteratively.

Another observation is that Transformer-based methods generally suffer from the unaligned-length problem~\cite{yu2020towards}, which denotes that Transformer is hard to correct the vision prediction if the predicted text length is unaligned with the ground truth. The unaligned-length problem is caused by the inevitable implementation that padding mask is fixed for filtering context outside the text length. Our iterative LM can alleviate this problem as the visual features and linguistic features are fused several times, and thus the predicted text length is also refined gradually.

\subsection{Fusion Model}
\label{sec:fusion}

Conceptually, the VM trained using images and the LM trained using text come from different modalities. To align visual features and linguistic features, we simply use the gated mechanism \cite{yu2020towards, yue2020robustscanner} for the final decision:
\begin{align}
\mathbf{G} &= \sigma([\mathbf{F}_{v}, \mathbf{F}_{l}] \mathbf{W}_f), \\
\mathbf{F}_{f} &= \mathbf{G} \odot \mathbf{F}_{v} + (1 - \mathbf{G}) \odot \mathbf{F}_{l},
\end{align}
where $\mathbf{W}_f \in \mathbb{R}^{2d \times d}$ is a parameter. $\mathbf{F}_{v}, \mathbf{F}_{l}, \mathbf{F}_{f}$ are vision, language, fusion features, respectively. $\sigma$ is a sigmoid function.  $\mathbf{G} \in \mathbb{R}^{t \times d}$ dynamically selects features from $\mathbf{F}_{v}$ and $\mathbf{F}_{l}$.

\subsection{Supervised Training}

ABINet++ is trained end-to-end with the following multi-task objectives:
\begin{align}
\mathcal{L} &= \lambda_v \mathcal{L}_v + \frac{\lambda_l}{M} \sum^M_{i=1}{\mathcal{L}^i_l} + \frac{1}{M} \sum^M_{i=1}{\mathcal{L}^i_f},
\label{eq:loss}
\end{align}
where $\mathcal{L}_v$, $\mathcal{L}_l$ and $\mathcal{L}_f$ are the cross entropy losses from $\mathbf{F}_{v}$, $\mathbf{F}_{l}$ and $\mathbf{F}_{f}$, respectively. Specifically, $\mathcal{L}^i_{l}$ and $\mathcal{L}^i_{f}$ are the losses at $i$-th iteration. $\lambda_v$ and $\lambda_l$ are balanced factors. $M$ is the iteration number.

\subsection{Semi-supervised Ensemble Self-training}
\label{sec:semi-supervised}

To further explore the superiority of our iterative model, we propose a semi-supervised learning method based on self-training~\cite{xie2020self} with the ensemble of iterative predictions. The basic idea of self-training is first to generate pseudo labels by the model itself, and then re-train the model using additional pseudo labels. Therefore, the key problem lies in constructing high-quality pseudo labels.

To filter the noise pseudo labels we propose the following methods: 1) minimum confidence of characters within a text instance is chosen as the text certainty. 2) Iterative predictions of each character are viewed as an ensemble to smooth the impact of noise labels. Therefore, we define a filtering function as follows:
\begin{align}
\begin{cases}
\mathcal{C} &= \min\limits_{1 \le t \le T} e^{\mathbb{E}[\log{P(y_t)}]} \\ 
P(y_t) &= \max\limits_{1 \le m \le M} P_m(y_t) 
\end{cases},
\label{eq:filter}
\end{align}
where $\mathcal{C}$ is the minimum \emph{certainty} of a text instance. $P_m(y_t)$ is the probability distribution of the $t$-th character at the $m$-th iteration. The training procedure is depicted in Algorithm~\ref{alg:self-training}, where $Q$ is a threshold. $B_l$, $B_u$ are training batches from labeled and unlabeled data. $N_{max}$ is the maximum number of training steps and $N_{upl}$ is the step number for updating pseudo labels.

\begin{algorithm}[t]
   \scriptsize
   \caption{Ensemble Self-training}
   \begin{algorithmic}[1]
      \Require Labeled images $\mathcal{X}$ with labels $\mathcal{Y}$ and unlabeled images $\mathcal{U}$
      \State Train parameters $\theta_0$ of ABINet++ with $(\mathcal{X}$, $\mathcal{Y})$ using Equation~\ref{eq:loss}.
      \State Use $\theta_0$ to generate soft pseudo labels $\mathcal{V}$ for $\mathcal{U}$
      \State Get $(\mathcal{U}'$, $\mathcal{V}')$ by filtering $(\mathcal{U}$, $\mathcal{V})$ with $\mathcal{C}<Q$ (Equation~\ref{eq:filter})
      \For{$i = 1,\ldots, N_{max}$}
         \If{$i == N_{upl}$}
            \State Update $\mathcal{V}$ using $\theta_i$
            \State Get $(\mathcal{U}'$, $\mathcal{V}')$ by filtering $(\mathcal{U}$, $\mathcal{V})$ with $\mathcal{C}<Q$ (Equation~\ref{eq:filter})
         \EndIf
         \State Sample $B_l=(\mathcal{X}_{b}$, $\mathcal{Y}_{b}) \subsetneqq (\mathcal{X}$, $\mathcal{Y})$, $B_u=(\mathcal{U}'_{b}$, $\mathcal{V}'_{b}) \subsetneqq (\mathcal{U}'$, $\mathcal{V}')$
         \State Update $\theta_i$ with $B_l$, $B_u$ using Equation~\ref{eq:loss}.
      \EndFor
   \end{algorithmic}
   \label{alg:self-training}
\end{algorithm}


\section{End-to-End Scene Text Spotting}

\begin{figure}
  \begin{center}
     \includegraphics[width=.5\textwidth]{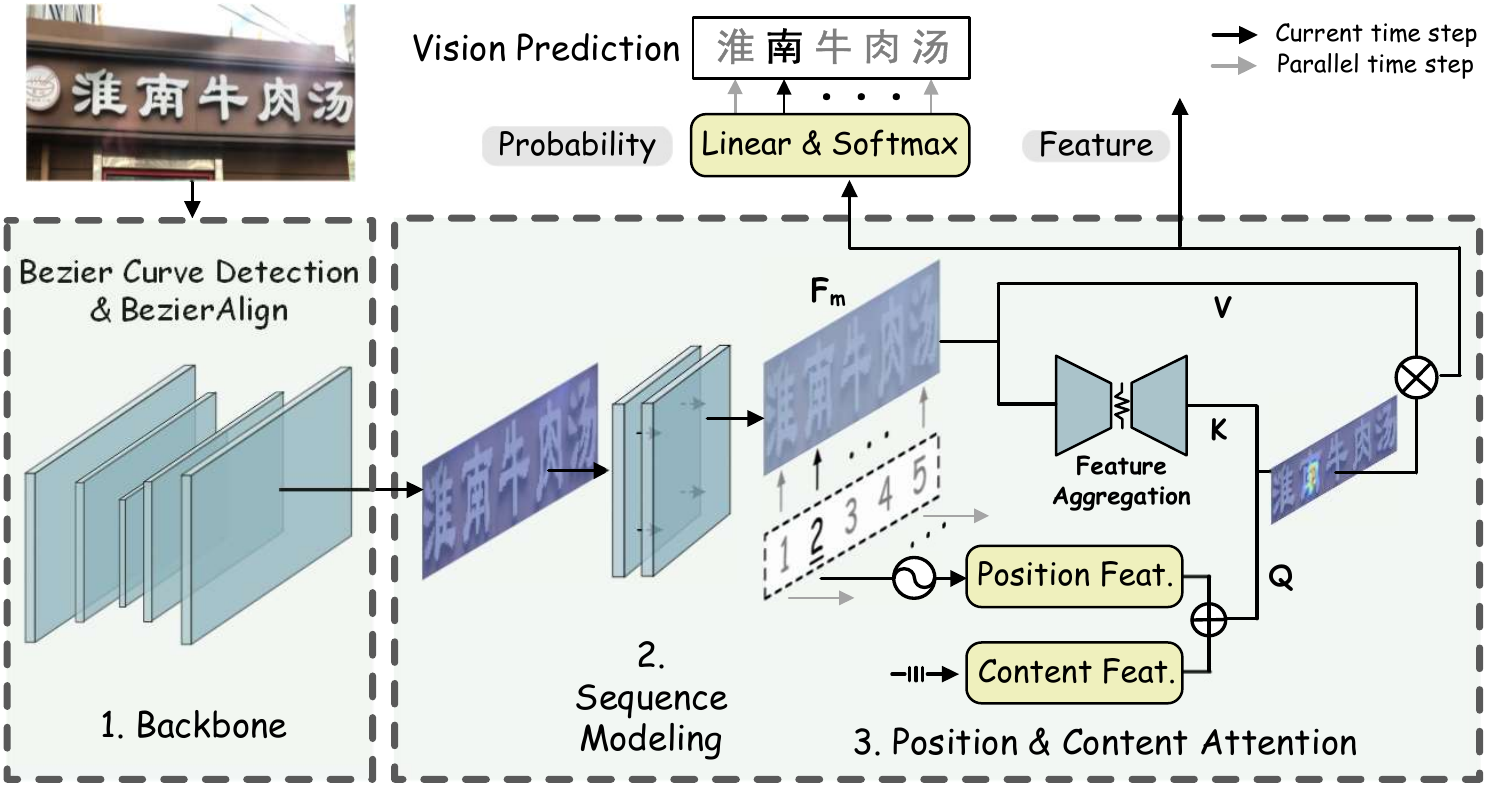}
     \caption{\textbf{Architecture of vision model for text spotting.}}
     \label{fig:vision-e2e}
  \end{center}
  \vspace{-1em}    
\end{figure}

\subsection{Vision Model for Text Spotting}

As shown in Fig.~\ref{fig:vision-e2e}, ABINet++ can be easily applied to end-to-end text spotting by extending the VM with a text detection module. The VM for text spotting inherits the main ideas from the VM for text recognition (Section~\ref{sec:vision}), where the attention mechanism is used for character decoding and Bezier curve detection~\cite{liu2020abcnet,liu2020abcnetv2} is introduced for text detection. Compared with the segmentation-based and other regression-based text detection methods, Bezier curve detection in ABCNet is naturally chosen as our text detection method as 1) it has a concise while effective pipeline without intricate post-processing. 2) The BezierAlign~\cite{liu2020abcnet} is able to transform multi-scale and arbitrarily-shaped text into normalized-size and horizontal-layout text features, which could be directly processed by our attention module in the same manner as in the text recognition (Section~\ref{sec:vision}).

Specifically, we apply BiFPN~\cite{tan2020efficientdet, liu2020abcnetv2} as our backbone. Based on the anchor-free object detection method FCOS~\cite{tian2019fcos}, eight control points for two Bezier curves (\ie, the 3rd-order Bezier curves) are regressed to form each text instance~\cite{liu2020abcnet}. Therefore, Equation~\ref{eq:vision} can be rewritten as follows:
\begin{equation}
\mathbf{F}_b = \mathcal{B}(\bm{\bar{x}}) \in \mathbb{R}^{k \times h \times w \times d}, 
\end{equation}
in which $\mathcal{B}(\cdot)$ is the Bezier curve detection model, and $k$ is the number of text instances within a full image $\bm{\bar{x}}$.

\subsection{Long Text Recognition}

\begin{figure}
   \begin{center}
      \includegraphics[width=0.5\textwidth]{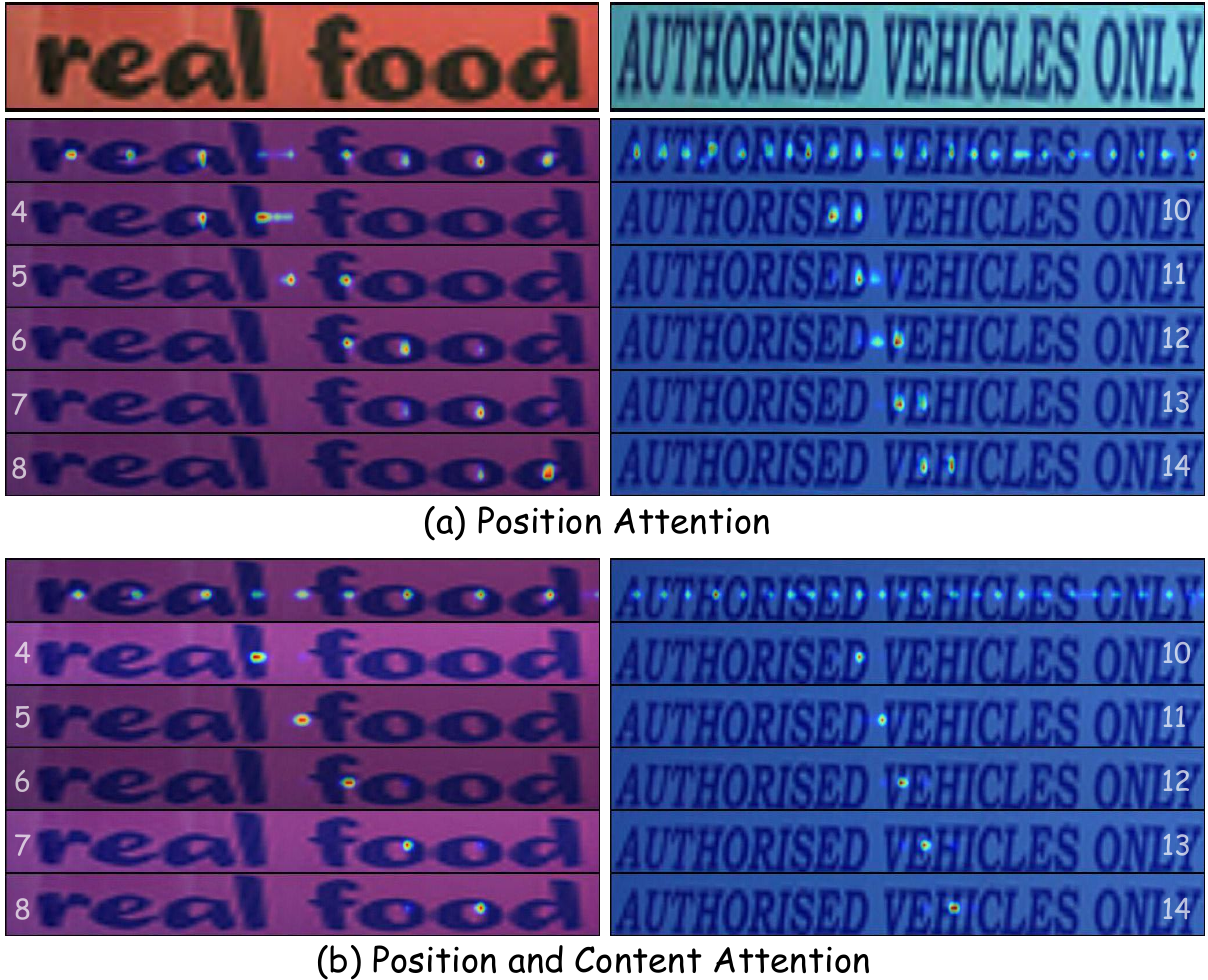}
      \caption{\textbf{Visualization of attention maps on SCUT-CTW1500 dataset.} Models are trained on SCUT-CTW1500 without pre-training, and BezierAlign is used to rectify curved text. Each row depicts a query step. Only truncated steps with obvious multi-activation phenomenon are shown.}
      \label{fig:long-text}
   \end{center}
   \vspace{-1.5em}     
\end{figure}

To decouple linguistic context from the VM, our conference version~\cite{fang2021read} adopts a concise attention strategy based on position query. Despite its effectiveness in common scenes, we observe the position-based attention~\cite{fang2021read,yu2020towards,lyu20192d} confronts performance degradation both in accuracy and speed when applied on long text images (e.g., SCUT-CTW1500 dataset with up to 100+ characters for a text instance). As presented in Fig.~\ref{fig:long-text}(a), there occurs a frequent multi-activation phenomenon under the setting for long text recognition, where a position query attains multiple response centers discretely on the attention maps. For example, the 4th position in ``real food" gets high responses on the characters ``a" and ``l" simultaneously. The multi-activation phenomenon suggests that the position-based attention can retrieve the character locations within a text instance. However, without the consideration of character content, the attention module becomes less confident to determine which response center the current query falls into, especially when dealing with long text images. To improve the performance of ABINet++ in long text recognition, we propose the following strategies:

\textbf{Convolution Sequence Modeling (CSM)}. In our scene text recognition version~\cite{fang2021read}, Transformer is employed as the SMN to capture long-range dependency on spatial features. For long text recognition, it is necessary to have a bigger size for spatial features, \eg, $8 \times 128$ versus $8 \times 32$ compared with regular-length text ~\cite{liu2020abcnetv2}. However, Transformer is inapplicable for long sequences due to the limitation of quadratic time complexity and high memory usage~\cite{zhou2021informer}. Thus for scene text spotting, we replace the Transformer units with stacked convolution layers as the SMN after BezierAlign. We discuss the impact in Section~\ref{sec:exp_e2e_ablation}.

\textbf{Horizontal Feature Aggregation (HFA)}. As BezierAlign transforms arbitrarily-shaped text into horizontal-representation text, we can assume that our recognition model processes characters in a horizontal arrangement. To suppress the multi-activation phenomenon, we experimentally explore some feature representation methods, and find that it is important to enhance information flow along the direction of character arrangement (Section~\ref{sec:exp_e2e_long_text}). Notice that an additional U-Net can aggregate information as the downsampling operation enlarges the receptive fields. With the same consideration, we propose to enhance U-Net structure by embedding a Transformer network after the finest feature map (Table~\ref{tab:unet}), in which only the horizontal features are retained and interacted with each other by a non-local mechanism.

\textbf{Position and Content Attention (PCA)}. To exclude multiple response centers, the discrimination of query vector is enhanced by integrating character content feature. We turn to the iterative strategy again (Section~\ref{sec:iterative}) to compute an accurate content feature. Specifically, for the initial iteration, a coarse content vector is directly predicted based on $\mathbf{F}_{m}$. For the subsequent iterations, visual feature $\mathbf{F}_{v}$ is reused as the content vector and refined with increasing iterations. Thus Equation~\ref{eq:query} can be rewritten as:
\begin{align}
\mathbf{Q} &= PE(t) + \mathcal{C}^{i}  \in \mathbb{R}^{t \times d}, \\
\mathcal{C}^{i} &= \begin{cases} \mathbf{W}_c \frac{\sum_{h,w}{\mathbf{F}_{m}}}{h*w}, & i = 0 \\ \mathbf{F}_{v}^{i-1}, & i > 0 \end{cases}, 
\end{align}
where $\mathbf{W}_c \in \mathbb{R}^{t \times d}$. $i$ is the iteration index and we empirically set the iteration number to 3. Note that all the content vectors are learned under the supervision of ground-truth labels. Therefore, the training loss of the VM for text spotting is:
\begin{align}
\mathcal{L}_v &= \sum^3_{i=1}{\mathcal{L}_{\mathcal{C}^{i}}} + \mathcal{L}_{det},
\end{align}
where $\mathcal{L}_{\mathcal{C}^{i}}$ is the cross entropy based on $\mathcal{C}^{i}$ and $\mathcal{L}_{det}$ is the detection loss proposed in~\cite{liu2020abcnetv2,liu2020abcnet}.

\subsection{Text Augmentation for Language Model}
\label{sec:text-aug}

\indent\textbf{Online Sampling Augmentation (OSA)}. Typically, training an LM requires more data than a VM~\cite{radford2018improving,qin2019towards}. However, when training a text spotter end-to-end, the data amount for the LM depends on the number of ground-truth/predicted instances in training images, which results in limited training samples. Besides, for the same reason, the batch size of the LM varies in different image batches, which would slow down the converging rate if the number of ground-truth/predicted instances is small. Therefore, we propose an online sampling method for text augmentation. Specifically, the batch size of the LM is fixed to $B_l$. During end-to-end training we randomly sample $B_s$ text instances from an additional text dataset, in which $B_s = max(0, B_l-B_o)$ and $B_o$ is the original batch size for LM. Note that the $B_l$ text instances are randomly chosen if $B_l < B_o$. This simple yet effective method enables our end-to-end model to learn from an external dataset, and achieve faster learning with constant batch size. In addition, OSA improves the utilization rate of GPU memory since the batch size of the LM is controllable.

\indent\textbf{Spelling Alteration Augmentation (SAA)}. During end-to-end training, OSA samples a text instance $\mathbf{t}$ as both the input and target for the LM (\ie, self-supervision). However, there is a discrepancy between the data distribution in the training and inference stages, as the input to the inference model may contain spelling errors. Therefore, we propose to simulate the noise input from the sampled text and attempt to reconstruct the ground-truth target. Specifically, for a sampled text $\mathbf{t}$, we either randomly modify one character by selecting one operation from \emph{replace}, \emph{insert} and \emph{delete} with the probabilities of $p_r$, $p_i$ and $p_d$, respectively, or remain the text unchanged with the probability of $p_u$, where $p_r + p_i + p_d + p_u = 1$. By this way, we can obtain abundant and meaningful samples for training the LM online.


\section{Experiment on Scene Text Recognition}
\label{sec:exp}

In the purpose of a fair comparison, experiments are conducted following the setup of~\cite{yu2020towards} unless otherwise stated. Six standard benchmarks are used for evaluation, including regular datasets ICDAR 2013~\cite{karatzas2013icdar}, Street View Text~\cite{wang2011end}, IIIT 5K-Words~\cite{mishra2012scene} and irregular datasets ICDAR 2015~\cite{karatzas2015icdar}, Street View Text-Perspective~\cite{quy2013recognizing}, CUTE80~\cite{risnumawan2014robust}. The regular datasets aim to evaluate the performance on horizontal characters, and the irregular datasets typically focus on non-horizontal text, \eg, curved, rotated and distorted text~\cite{baek2019wrong}.

\subsection{Implementation Details}

The supervised training is performed using two synthetic datasets MJSynth~\cite{jaderberg2014synthetic,jaderberg2016reading} and SynthText~\cite{gupta2016synthetic} without fine-tuning on real training datasets. Uber-Text~\cite{Ying2017UberText} is chosen as an unlabeled dataset to evaluate the semi-supervised method. The model dimension $d$ for the recognition model is set to 512 throughout. There are 4 layers in BCN with 8 attention heads each layer. The balanced factors $\lambda_v$, $\lambda_l$ are set to 1, 1 respectively. Images are directly resized to $32 \times 128$ with data augmentation such as geometry transformation (\ie, rotation, affine and perspective), image quality deterioration and color jitter, \etc. The model recognizes 36 character classes, including 10 digits and 26 case-insensitive letters. The max length of output text is set to 25. We use 4 NVIDIA 1080Ti GPUs to train our recognition models with batch size 384. Adam optimizer is adopted with the initial learning rate $1e^{-4}$, which is decayed to $1e^{-5}$ after 6 epochs.

\subsection{Datasets}

\textbf{ICDAR 2013 (IC13)}~\cite{karatzas2013icdar} has two versions for evaluation, \ie, the 1,015 image version and 857 image version. \textbf{Street View Text (SVT)}~\cite{wang2011end} contains 647 testing images. \textbf{IIIT 5K-Words (IIIT)}~\cite{mishra2012scene} includes 3,000 images for evaluation. \textbf{ICDAR 2015 (IC15)}~\cite{karatzas2015icdar} contains 2,077 testing images, and among them 1,811 images are commonly kept for evaluation. The number of evaluation images in \textbf{Street View Text-Perspective (SVTP)}~\cite{quy2013recognizing} is 645. \textbf{CUTE80 (CUTE)}~\cite{risnumawan2014robust} consists of 288 word-box images for evaluation. \textbf{MJSynth (MJ)} is a synthetic dataset~\cite{jaderberg2014synthetic,jaderberg2016reading} including about 8.92M word-box images for training. \textbf{SynthText (ST)}~\cite{gupta2016synthetic} has about 6.98M word-box images for training, which are cropped from synthetic images. \textbf{Uber-Text}~\cite{Ying2017UberText} contains 571,534 word-box images cropped from scene images. Detailed descriptions of these datasets are given in the supplementary material.

\subsection{Ablation Study}
\subsubsection{Vision Model}

\begin{table}
   \begin{center}
   \caption{Ablation study of the vision model. “Attn" is the attention method. “Trans. Layer" is the number of Transformer layers. “SV", “MV$_{\text{pos}}$", “MV$_{\text{par}}$", “LV" are the tags for four vision models, differing in attention method and Transformer layers.}
   \label{tab:vision}
	\resizebox{1.0\linewidth}{!}{
   \begin{tabular}{c|c|c|c|c|c|c|c|c}
      \whline
      Model & \multirow{2}{*}{Attn} & Trans. & IC13 & SVT & IIIT & \multirow{2}{*}{Avg} & Params & Time\tablefootnote{The inference time in Table~\ref{tab:vision},~\ref{tab:bidirectional},~\ref{tab:iterative},~\ref{tab:benchmark} is estimated using an NVIDIA Tesla V100 by averaging 3 different trials.} \\
      Name &  & Layer & IC15 & SVTP & CUTE & & ($\times10^6$) & (ms) \\
      \hline
      SV & \multirow{2}{*}{parallel} & \multirow{2}{*}{2} &94.2& 89.6 & 93.7& \multirow{2}{*}{88.8} & \multirow{2}{*}{19.6} & \multirow{2}{*}{12.5} \\
      (small) & & & 80.6& 82.3& 85.1&  & & \\ 
      \hline
      MV$_{\text{pos}}$ & \multirow{2}{*}{position} & \multirow{2}{*}{2} &93.6& 89.3 & 94.2& \multirow{2}{*}{89.0} & \multirow{2}{*}{20.4}  & \multirow{2}{*}{14.9} \\
      (middle) & & & 80.8& 83.1& 85.4&  &  & \\
      \hline
      MV$_{\text{par}}$ & \multirow{2}{*}{parallel} & \multirow{2}{*}{3} &94.5& 89.5 & 94.3& \multirow{2}{*}{89.4} & \multirow{2}{*}{22.8} & \multirow{2}{*}{14.8}  \\
      (middle) & & & 81.1& 83.7& \bf{86.8}&  & & \\  
      \hline
      LV & \multirow{2}{*}{position} & \multirow{2}{*}{3} &\bf{94.9}& \bf{90.4} & \bf{94.6}& \multirow{2}{*}{\bf{89.8}} & \multirow{2}{*}{23.5} & \multirow{2}{*}{16.7}  \\
      (large) & & & \bf{81.7} & \bf{84.2} & 86.5&  & & \\  
      \whline
   \end{tabular}}
   \end{center}
   \vspace{-2em} 
\end{table}

\begin{figure}
   \begin{center}
      \includegraphics[width=0.5\textwidth]{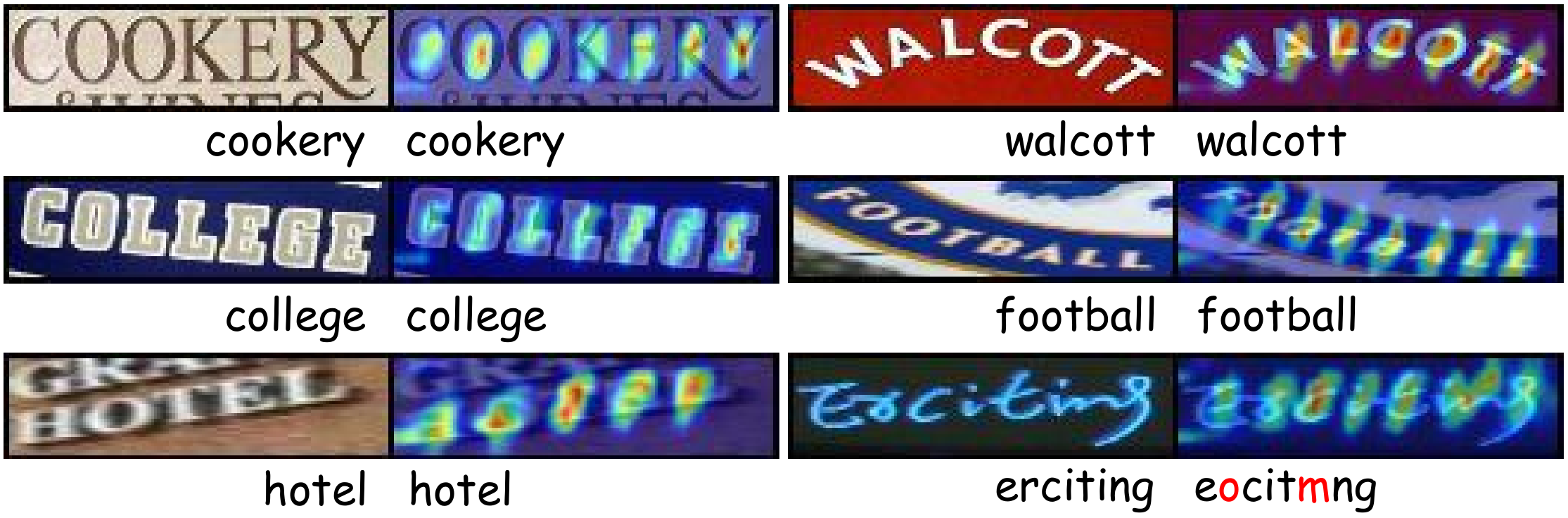}
      \caption{\textbf{Visualization of attention maps for vision model.} Text strings are ground truth and vision prediction, respectively.}
      \label{fig:attention_map}
   \end{center}
   \vspace{-0.5em}   
\end{figure}

Firstly, we discuss the performance of VM from two aspects: the character prediction and the sequence modeling. Experiment results are recorded in Table~\ref{tab:vision}. The \emph{parallel} attention is a popular attention method~\cite{lyu20192d,yu2020towards}, and the proposed \emph{position} attention has a more powerful representation for the key/value vectors. From the statistics we can conclude: 1) simply upgrading the VM will result in great gains in accuracy but at the cost of the parameter and speed. 2) To upgrade the VM, we can use the position attention in the character prediction stage and a deeper Transformer in the sequence modeling stage. Visualization of the attention maps for LV model is given in Fig.~\ref{fig:attention_map}.

\subsubsection{Language Model}

\begin{table}
   \begin{center}
   \caption{Ablation study of the autonomous strategy. “PVM" is pre-training the vision model on MJ and ST. “PLM$_{in}$" is pre-training the language model using the text on MJ and ST. “PLM$_{out}$" is pre-training the language model on WikiText-103~\cite{merity2016pointer}. “AGF" means allowing gradient flow between the vision model and the language model.}
   \label{tab:autonomous}
	\resizebox{.9\linewidth}{!}{
   \begin{tabular}{c|c|c|c|c|c|c|c}
      \whline
      \multirow{2}{*}{PVM} & \multirow{2}{*}{PLM$_{in}$} & \multirow{2}{*}{PLM$_{out}$} & \multirow{2}{*}{AGF} & IC13 & SVT & IIIT & \multirow{2}{*}{Avg}  \\
       & & & & IC15 & SVTP & CUTE &  \\
      \hline
      \multirow{2}{*}{-} & \multirow{2}{*}{-} & \multirow{2}{*}{-} & \multirow{2}{*}{-} & 96.7 & 93.4 & 95.7 & \multirow{2}{*}{91.7}  \\
      & & & & 84.5 & 86.8 & 86.8&   \\ 
      \hline
      \multirow{2}{*}{\checkmark} & \multirow{2}{*}{-} & \multirow{2}{*}{-} & \multirow{2}{*}{-} & 97.0 & 93.0 & 96.3 & \multirow{2}{*}{92.3}  \\
       & & & & 85.0 & 88.5 & 89.2&  \\ 
       \hline
      \multirow{2}{*}{-} & \multirow{2}{*}{\checkmark} & \multirow{2}{*}{-} & \multirow{2}{*}{-} & 97.1 & \bf{93.8} & 95.5 & \multirow{2}{*}{91.6}  \\
       & & & & 83.6 & 88.1 & 86.8&  \\ 
      \hline
      \multirow{2}{*}{\checkmark} & \multirow{2}{*}{\checkmark} & \multirow{2}{*}{-} & \multirow{2}{*}{-} & \bf{97.2} & 93.5 & 96.3 & \multirow{2}{*}{92.3}  \\
      & & & & 84.9 & \bf{89.0} & 88.5&  \\ 
      \hline
      \multirow{2}{*}{\checkmark} & \multirow{2}{*}{-} & \multirow{2}{*}{\checkmark} & \multirow{2}{*}{-} & 97.0 & 93.7 & \bf{96.5} & \multirow{2}{*}{\bf{92.5}}  \\
      & & & & \bf{85.3} & 88.5 & \bf{89.6}&  \\ 
      \hline
      \multirow{2}{*}{\checkmark} & \multirow{2}{*}{-} & \multirow{2}{*}{-} & \multirow{2}{*}{\checkmark} & 96.7 & 92.6 & 95.7& \multirow{2}{*}{91.4}  \\
      & & & & 83.3 & 86.5 & 88.5&  \\ 
      \whline
   \end{tabular}}
   \end{center}
   \vspace{-0.5em} 
\end{table}

\textbf{Autonomous Strategy.} To analyze the autonomous models, we adopt the LV and BCN as VM and LM, respectively. As the VM and LM can be viewed as independent models, both of them can be easily pre-trained and then used to initialize the full ABINet++ model. The VM is pre-trained on the images of MJ and ST in a supervised way. The LM can be pre-trained based on raw text using self-supervised learning, where the input text is either from the annotation text in MJ and ST, or from an external unlabeled dataset (\eg, WikiText-103). From the results in Table~\ref{tab:autonomous} we can observe: 1) pre-training VM is useful which boosts the accuracy about $0.6\%$-$0.7\%$ on average; 2) the benefit of pre-training LM on the training datasets (\ie, MJ and ST) is negligible; 3) while pre-training LM from an additional unlabeled dataset (\eg, WikiText-103) is helpful even when the base model is in high accuracy. The above observations suggest that it is useful for scene text recognition to pre-train both VM and LM. Pre-training LM on additional unlabeled datasets is more effective than on the training datasets since the limited text diversity and biased data distribution are unable to facilitate the learning of a well-performed LM. Also, pre-training LM on unlabeled datasets is cheap since additional data is available easily.

Besides, by allowing gradient flow (AGF) between VM and LM, the performance decreases by $0.9\%$ on average (Table~\ref{tab:autonomous}). We also notice that the training loss of AGF reduces sharply to a lower value. This indicates that overfitting occurs in the LM as the VM helps to cheat in training, which might also happen in implicit language modeling. Therefore it is crucial to enforce LM to learn independently by BGF. We note that SRN~\cite{yu2020towards} uses a \emph{argmax} operation after the VM, which is intrinsically a special case of BGF since \emph{argmax} is non-differentiable. Another advantage is that the autonomous strategy allows better interpretability, since we can have a deep insight into the performance of LM (\eg, Table~\ref{tab:spelling_correction}), which is infeasible in implicit language modeling.

\begin{table}
   \begin{center}
   \caption{\textls[-13]{Ablation study of the bidirectional representation. “SRN-U" is a unidirectional version of SRN. “BERT-E" is a BERT-based language model trying to correct erroneous characters only.}}
   \label{tab:bidirectional}
	\resizebox{1.\linewidth}{!}{
   \begin{tabular}{c|c|c|c|c|c|c|c|c}
      \whline
      \multirow{2}{*}{Vision} & \multirow{2}{*}{Language} & IC13 & SVT & IIIT & \multirow{2}{*}{Avg} & Params & FLOPs & Time \\
       &  & IC15 & SVTP & CUTE & & ($\times10^6$) & ($\times10^9$) & (ms) \\
      \hline
       & \multirow{2}{*}{SRN-U} & 96.0 & 90.3 & 94.9& \multirow{2}{*}{90.2} & \multirow{2}{*}{32.8} & \multirow{2}{*}{9.76}  & \multirow{2}{*}{19.1} \\
       & & 81.9 & 86.0 & 85.4&  & & \\ 
      \cline{2-9}
      \multirow{2}{*}{SV} & \multirow{2}{*}{SRN} & 96.3 & 90.9 & 95.0 & \multirow{2}{*}{90.6} & \multirow{2}{*}{45.4} & \multirow{2}{*}{10.18} & \multirow{2}{*}{24.2} \\
      & & 82.6 & 86.4 & 87.5&  & & \\ 
      \cline{2-9}
      & \multirow{2}{*}{BERT-E} & 96.4 & 91.2 & 95.2 & \multirow{2}{*}{90.7} & \multirow{2}{*}{32.8} & \multirow{2}{*}{--}  & \multirow{2}{*}{29.1} \\
      & & 83.4 & 84.0 & 87.8 &  & & \\ 
      \cline{2-9}
      & \multirow{2}{*}{BERT} & 96.7 & 92.0 & 95.1 & \multirow{2}{*}{91.0} & \multirow{2}{*}{32.8} & \multirow{2}{*}{20.66}  & \multirow{2}{*}{38.0} \\
      & & 83.4 & 85.4 & 88.2 &  & & \\
      \cline{2-9}
      & \multirow{2}{*}{BCN} & 96.7 & 91.7 & 95.3 & \multirow{2}{*}{91.0} & \multirow{2}{*}{32.8} & \multirow{2}{*}{9.76}  & \multirow{2}{*}{19.5} \\
      & & 83.1 & 86.2 & 88.9 &  & & \\       
      \hline
      \hline
      & \multirow{2}{*}{SRN-U} & 96.0 & 91.2 & 96.2 & \multirow{2}{*}{91.5} & \multirow{2}{*}{36.7} & \multirow{2}{*}{10.94}  & \multirow{2}{*}{22.1} \\
      & & 84.0 & 86.8 & 87.8&  & & \\ 
      \cline{2-9}
      \multirow{2}{*}{LV} & \multirow{2}{*}{SRN} & 96.8 & 92.3 & 96.3 & \multirow{2}{*}{91.9} & \multirow{2}{*}{49.3} & \multirow{2}{*}{11.36}  & \multirow{2}{*}{26.9} \\
      & & 84.2 & 87.9 & 88.2&  & & \\ 
      \cline{2-9}
      & \multirow{2}{*}{BERT-E} & 95.6 & 92.6 & 95.7 & \multirow{2}{*}{91.8} & \multirow{2}{*}{36.7} & \multirow{2}{*}{--}  & \multirow{2}{*}{32.2} \\
      & & 85.1 & 87.6 & 89.9 &  & & \\ 
      \cline{2-9}
      & \multirow{2}{*}{BERT} & 96.9 & 93.9 & 96.3 & \multirow{2}{*}{92.3} & \multirow{2}{*}{36.7} & \multirow{2}{*}{21.84}  & \multirow{2}{*}{41.3} \\
      & & 84.8 & 88.4 & 89.5 &  & & \\ 
      \cline{2-9}
      & \multirow{2}{*}{BCN} & 97.0 & 93.0 & 96.3 & \multirow{2}{*}{92.3} & \multirow{2}{*}{36.7} & \multirow{2}{*}{10.94}  & \multirow{2}{*}{22.0} \\
      & & 85.0 & 88.5 & 89.2&  & & \\             
      \whline
   \end{tabular}}
   \end{center}
   \vspace{-0.5em}   
\end{table}

\begin{table}
   \begin{center}
   \caption{Top-5 accuracy of language models in text spelling correction.}
   \resizebox{0.85\linewidth}{!}{
   \begin{tabular}{c|c|c}
      \whline
      Language Model & Character Accuracy & Word Accuracy \\
      \hline
      SRN & 78.3 & 27.6 \\
      \hline
      BCN & \bf{82.8} & \bf{41.9} \\
      \whline
   \end{tabular}}
   \label{tab:spelling_correction}
   \end{center}
   \vspace{-1.5em}
\end{table}

\textbf{Bidirectional Representation.} As BCN is a variant of Transformer, we compare BCN with its counterpart SRN. The Transformer-based SRN~\cite{yu2020towards} shows superior performance which is an ensemble of unidirectional representation. For a fair comparison, experiments are conducted under the same conditions except for the networks. We use SV and LV as the VMs to validate the effectiveness at different accuracy levels. As depicted in Table~\ref{tab:bidirectional}, though BCN has similar parameters and inference speed as the unidirectional version of SRN (SRN-U), it achieves a competitive advantage in accuracy under different VMs. Besides, compared with the bidirectional SRN in an ensemble, BCN shows better performance especially on challenging datasets such as IC15 and CUTE. Also, ABINet++ equipped with BCN is about $20\%$-$25\%$ faster than SRN, which is practical for large-scale tasks.

To comprehensively compare different LMs, we also introduce a BERT-based correction model similar to~\cite{bhunia2021joint}. For each character $y_i$ in a text instance $\bm{y}_{1:n}$ predicted by the VM, this model repeatedly replaces $y_i$ with token {\tt{[MASK]}} and predicts the masked token, necessitating $n$ separate BERT passes. To reduce the computation complexity, we attempt to filter correctly predicted characters using a selected confidence threshold, therefore only the erroneous characters are predicted and unnecessary computation can be avoided. For a fair comparison with BCN, this BERT is implemented by a 4-layer Transformer encoder. Analyzing the results in Table~\ref{tab:bidirectional} we can know, the BERT shows comparable recognition accuracy to our BCN. However, the BERT-based method is of low efficiency since the repeated computation requires a large number of FLOPs and GPU memory, while trying to reduce computation complexity may harm its accuracy (\ie, “BERT-E" in Table~\ref{tab:bidirectional}). Also note that the above experiments set $n$ to regular length 25, it becomes worse for the BERT-based method with an increasing text length.

Section~\ref{sec:autonomous} has argued that the LMs can be viewed as independent units to estimate the probability distribution of spelling correction, and thus we conduct experiments from this view. The training set is the text from MJ and ST. To simulate spelling errors, the testing set is 20000 items that are chosen randomly, where we add or remove a character for $20\%$ text, replace a character for $60\%$ text and keep the rest of the text unchangeable. From the results in Table~\ref{tab:spelling_correction}, we can see BCN outperforms SRN by $4.5\%$ in character accuracy and $14.3\%$ in word accuracy, which indicates that BCN has a more powerful ability at character-level language modeling.

\begin{figure}
   \begin{center}
      \includegraphics[width=0.48\textwidth]{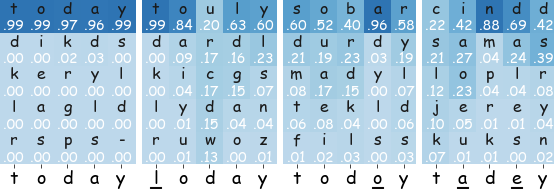}
      \caption{\textbf{Visualization of top-5 probability in BCN.}}
      \label{fig:visual_topk}
   \end{center}
   \vspace{0em}   
\end{figure}

To better understand how BCN works inside ABINet++, we visualize the top-5 probability in Fig.~\ref{fig:visual_topk}, which takes “today" as an example. On the one hand, as ``today" is a string with semantic information, taking ``-oday" and ``tod-y" as inputs, BCN can predict ``t" and ``a" with high confidence and contribute to final fusion predictions. On the other hand, as error characters ``l" and ``o" are noise for the rest predictions, BCN becomes less confident and has little impact on the final predictions. Besides, if there are multiple error characters, it is hard for BCN to restore correct text due to lacking enough context.

\begin{table}
   \begin{center}
   \caption{Ablation study of the iterative correction.}
   \label{tab:iterative}
	\resizebox{1.\linewidth}{!}{
   \begin{tabular}{c|c|c|c|c|c|c|c|c}
      \whline
      \multirow{2}{*}{Model} & Iteration & IC13 & SVT & IIIT & \multirow{2}{*}{Avg} & Params & FLOPs & Time \\
       & Number & IC15 & SVTP & CUTE & & ($\times10^6$) & ($\times10^9$) & (ms) \\
      \hline
      \multirow{2}{*}{SV} & \multirow{2}{*}{1} & 96.7 & 91.7 & 95.3 & \multirow{2}{*}{91.0} & \multirow{2}{*}{32.8} & \multirow{2}{*}{9.76} & \multirow{2}{*}{19.5} \\
      &  & 83.1 & 86.2  & 88.9 &  & & \\ 
      \cline{2-9}
      \multirow{2}{*}{+} & \multirow{2}{*}{2} & \bf{97.2} & 91.8 & \bf{95.4} & \multirow{2}{*}{91.2} & \multirow{2}{*}{32.8} & \multirow{2}{*}{10.18} & \multirow{2}{*}{24.5} \\
      &  & 83.3 & 86.4 & 89.2 &  & & \\ 
      \cline{2-9}
      \multirow{2}{*}{BCN} & \multirow{2}{*}{3}  & 97.1 & \bf{93.0} & \bf{95.4} & \multirow{2}{*}{\bf{91.4}} & \multirow{2}{*}{32.8} & \multirow{2}{*}{10.6} & \multirow{2}{*}{31.6} \\
      & & \bf{83.4} & \bf{86.7}  & \bf{89.6}&  & & \\ 
      \hline
      \hline
      \multirow{2}{*}{LV} & \multirow{2}{*}{1} & 97.0 & 93.0 & 96.3 & \multirow{2}{*}{92.3} & \multirow{2}{*}{36.7} & \multirow{2}{*}{10.94} & \multirow{2}{*}{22.0} \\
      &  & 85.0 & 88.5 & 89.2&  & & \\ 
      \cline{2-9}
      \multirow{2}{*}{+} & \multirow{2}{*}{2} & 97.1 & 93.4 & 96.3 & \multirow{2}{*}{92.4} & \multirow{2}{*}{36.7} & \multirow{2}{*}{11.36} & \multirow{2}{*}{27.3} \\
      &  & 85.2 & 88.7 & \bf{89.6}&  & & \\ 
      \cline{2-9}
      \multirow{2}{*}{BCN} & \multirow{2}{*}{3} & \bf{97.3} & \bf{94.0} & \bf{96.4} & \multirow{2}{*}{\bf{92.6}} & \multirow{2}{*}{36.7} & \multirow{2}{*}{11.78} & \multirow{2}{*}{33.9} \\
      &  & \bf{85.5} & \bf{89.1}  & 89.2 &  & & \\ 
      \whline
   \end{tabular}}
   \end{center}
   \vspace{-1.5em}
\end{table}

\begin{table*}[htp]
   \begin{center}
   \caption{Accuracy comparison with other methods for scene text recognition. “SV" and “LV" denote small and large vision models. $^*$ indicates reproduced version. $^\dag$ and $^\ddag$ are trained with self-training and ensemble self-training, respectively.}
   \newcommand{\tabincell}[2]{\begin{tabular}{@{}#1@{}}#2\end{tabular}}
   \label{tab:benchmark}
   \resizebox{1.\linewidth}{!}{
   \begin{tabular}{r|c|c|c|c|c|c|c|c|c|c|c|c}
   \whline
   \multirow{3}{*}{Methods} & \multirow{3}{*}{Year} & \multirow{3}{*}{\tabincell{c}{Labeled \\ Datasets}} & \multirow{3}{*}{\tabincell{c}{Unlabeled \\ Datasets}} & \multicolumn{4}{|c|}{Regular Text}  & \multicolumn{4}{|c|}{Irregular Text} & \multirow{3}{*}{\tabincell{c}{Time \\ (ms)}}\\
   \cline{5-12}
   & & & & IC13 & IC13 & SVT & IIIT & IC15 & IC15 & SVTP & CUTE & \\
   & & & & 857 & 1015 & 647 & 3000 & 1811 & 2077 & 645 & 288 & \\
   \hline
   SEED~Qiao~\etal~\cite{qiao2020seed} & 2020 & MJ+ST & - & - & 92.8 & 89.6 & 93.8 & 80.0 & - & 81.4 & 83.6 & 53.3 \\
   Textscanner~Wan~\etal~\cite{wan2019textscanner} & 2020 & MJ+ST & - & - & 92.9 & 90.1 & 93.9 & 79.4 & - & 84.3 & 83.3 & - \\
   DAN~Wang~\etal~\cite{wang2020decoupled} & 2020 & MJ+ST & - & - & 93.9 & 89.2 & 94.3 & - & 74.5 & 80.0 & 84.4 & - \\
   RobustScanner~Yue~\etal~\cite{yue2020robustscanner} & 2020 & MJ+ST & - & - & 94.8 & 88.1 & 95.3 & - & 77.1 & 79.5 & \bf{90.3} & 36.0 \\
   SRN~Yu~\etal~\cite{yu2020towards} & 2020 & MJ+ST & - & 95.5 & - & 91.5 & 94.8 & 82.7 & - &85.1 & 87.8 & 46.2\\
   PIMNet~Qiao~\etal~\cite{qiao2021pimnet} & 2021 & MJ+ST & - & 95.2 & 93.4 & 91.2 & 95.2 & 83.5 & 81.0 & 84.3 & 84.4 & 19.1 \\
   Bhunia~\etal~\cite{bhunia2021joint} & 2021 & MJ+ST & - & - & 95.5 & 92.2 & 95.2 & - & 84.0 & 85.7 & 89.7 & - \\
   VisionLan~Wang~\etal~\cite{wang2021two} & 2021 & MJ+ST & - & 95.7 & - & 91.7 & 95.8 & 83.7 & - & 86.0 & 88.5 & 11.5\\
   \hline
   SRN$^*$ (SV) & &  MJ+ST & - & 96.3 & - & 90.9 & 95.0 & 82.6 & - & 86.4 & 87.5 & 24.2\\
   ABINet++ (SV) & & MJ+ST & - & 96.8 & - & 93.2 &  95.4 & 84.0 & - &  87.0 & 88.9 & 31.6 \\  
   SRN$^*$ (LV) & &  MJ+ST & - & 96.8 & - & 92.3 & \bf{96.3} & 84.2 & - & 87.9 & 88.2 & 26.9 \\
   ABINet++ (LV) & & MJ+ST & - & \bf{97.4} & \bf{95.7} & \bf{93.5} & 96.2 & \bf{86.0} & \bf{85.1} & \bf{89.3}  & 89.2 & 33.9 \\
   \hline
   ABINet++$^\dag$ (LV) & & MJ+ST & Uber-Text & 97.3 & - & 94.9 & 96.8 & \bf{87.4} & - & \bf{90.1} & 93.4 & - \\
   ABINet++$^\ddag$ (LV) & & MJ+ST & Uber-Text & \bf{97.7} & - & \bf{95.5} & \bf{97.2} & 86.9 & - & 89.9  & \bf{94.1} & - \\
   \hline
   GTC~Hu~\etal~\cite{hu2020gtc} & 2020 & MJ+ST+Real & - & - & 94.4 & 92.9 & 95.8 & - & 79.5 & 85.7 & 92.2 & - \\
   Textscanner~Wan~\etal~\cite{wan2019textscanner} & 2020 & MJ+ST+Real & - & - & 94.9 & 92.7 & 95.7 & 83.5 & - & 84.8 & 91.6 & - \\
   PIMNet~Qiao~\etal~\cite{qiao2021pimnet} & 2021 & MJ+ST+Real & - & 96.6 & 95.4 & 94.7 & 96.7 & 88.7 & 85.9 & 88.2 & 92.7 & - \\
   ABINet++ (LV) & & MJ+ST+Real & - & \bf{98.1} & \bf{97.1} & \bf{96.1} & \bf{97.1} & \bf{89.2} & \bf{86.0} & \bf{92.2}  & \bf{94.4} & - \\
   \whline
   \end{tabular}}
   \end{center}
   \vspace{-1.5em}
 \end{table*}

\textbf{Iterative Correction.} We apply SV and LV again with BCN to demonstrate the performance of the iterative correction from different levels. Experiment results are given in Table~\ref{tab:iterative}, where the iteration numbers are set to 1, 2 and 3 both in training and testing. As can be seen from the results, iterating the BCN 3 times can respectively boost the accuracy by $0.4\%$, $0.3\%$ on average. Specifically, there is little gain on IIIT which is a relatively easy dataset with clear character appearance. However, when it comes to other harder datasets such as IC15, SVT and SVTP, the iterative correction steadily increases the accuracy and achieves up to $1.3\%$ and $1.0\%$ improvement on SVT for SV and LV respectively. It is also noted that the inference time increases linearly with the iteration number.

\begin{figure}
   \begin{center}
      \includegraphics[width=0.48\textwidth]{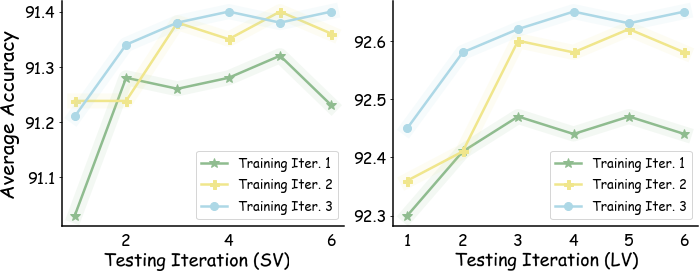}
      \caption{\textbf{Accuracy of iterating BCN in training and testing.} The vision models are SV (left) and LV (right), respectively.}
      \label{fig:iteration}
   \end{center}
   \vspace{-1.em} 
\end{figure}

We further explore the difference of the iteration correction between training and testing. The fluctuation of the average accuracy in Fig.~\ref{fig:iteration} suggests that: 1) directly applying the iterative correction in testing also works well; 2) iterating in training is further beneficial since it provides additional training samples for the LM; 3) the accuracy reaches a saturated state when iterating the model more than 3 times, and therefore a big iteration number is unnecessary.

\begin{figure}
   \begin{center}
      \includegraphics[width=0.5\textwidth]{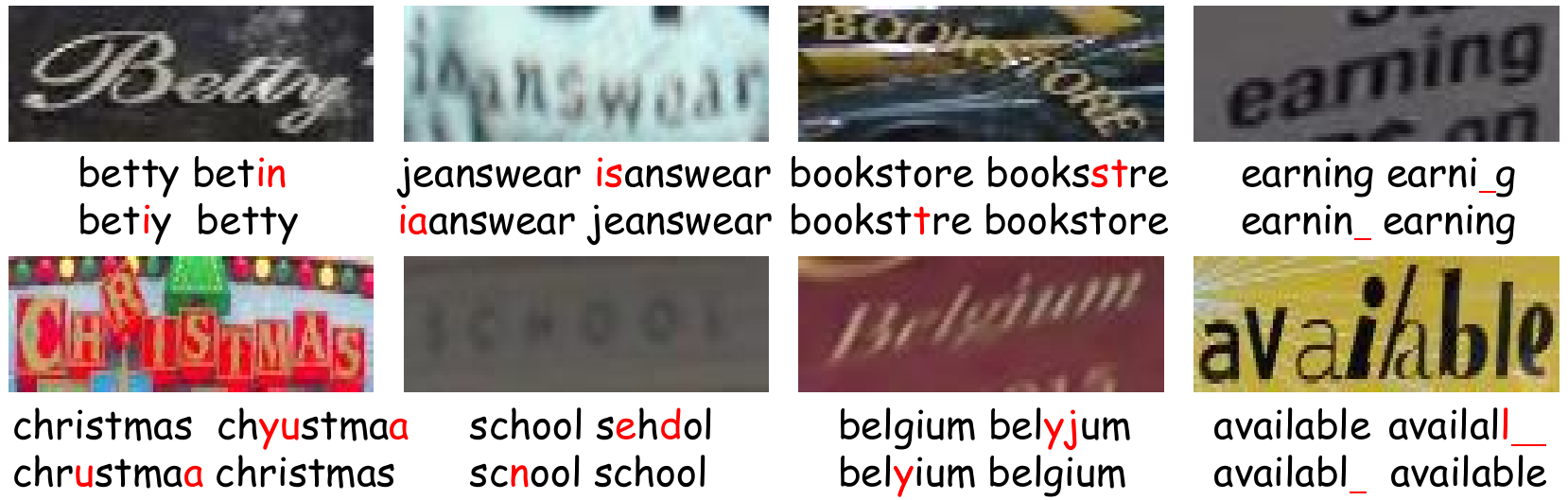}
      \caption{\textbf{Successful examples using the iterative correction.} Text strings are the ground truth, vision prediction, fusion prediction without the iterative correction and with the iterative correction respectively from left to right and top to bottom.}
      \label{fig:iteration_img}
   \end{center}
   \vspace{-1.5em}      
\end{figure}

To have a comprehensive cognition about the iterative correction, we visualize the intermediate predictions in Fig.~\ref{fig:iteration_img}. Typically, the vision predictions can be revised approaching to the ground truth, but in some cases the errors still might occur. After multiple iterations, the predictions can be corrected finally. Besides, we also observe that the iterative correction is able to alleviate the unaligned-length problem, as shown in the last column in Fig.~\ref{fig:iteration_img}. 

From the ablation study we can conclude: 1) the bidirectional BCN is a powerful LM which can effectively improve the performance both in accuracy and speed. 2) By further equipping BCN with the iterative correction, the noise input problem can be alleviated, which is recommended to deal with challenging examples such as low-quality images at the expense of incremental computations.

\subsection{Comparisons with State-of-the-Arts}

 Generally, it is not an easy job to fairly compare with other methods directly using the reported statistics~\cite{baek2019wrong}, as differences might exist in the backbone (\ie, CNN structure and parameters), data processing (\ie, images rectification and data augmentation) and training tricks, \etc. To strictly perform a fair comparison, we reproduce the state-of-the-art algorithm SRN which shares the same experiment configuration with ABINet++, as presented in Table~\ref{tab:benchmark}. The two reimplemented SRN (SV) and SRN (LV) are slightly different from the reported model by replacing VMs, removing the side-effect of multi-scale training, applying a decayed learning rate, etc. Note that SRN (SV) performs somewhat better than SRN due to the above tricks. As can be seen from the comparison, our ABINet++ (SV) outperforms SRN (SV) by $0.5\%$, $2.3\%$, $0.4\%$, $1.4\%$, $0.6\%$, $1.4\%$ on IC13, SVT, IIIT, IC15, SVTP and CUTE datasets respectively. Also, the ABINet++ (LV) with a stronger VM achieves an improvement of $0.6\%$, $1.2\%$, $1.8\%$, $1.4\%$, $1.0\%$ on IC13, SVT, IC15, SVTP and CUTE benchmarks over its counterpart.
 
 Compared with recent state-of-the-art works that are trained on MJ and ST, ABINet++ also shows impressive performance~(Table~\ref{tab:benchmark}). Especially, ABINet++ has prominent superiority on SVT, SVTP and IC15 as these datasets contain a large amount of low-quality images such as noise and blurred images, which the VM is not able to confidently recognize. Besides, we also find that images with unusual-font and irregular text can be successfully recognized as linguistic information acts as an important complement to visual features. Therefore ABINet++ can attain the second best result on CUTE even without image rectification.

To compare with the state-of-the-art methods trained on real data, we collect additional training data from public datasets following~\cite{qiao2021pimnet}. Concretely, we further fine-tune ABINet++ (LV) on a union set of IIIT5K, SVT, IC03, IC13, IC15 and COCO. As seen in Table~\ref{tab:benchmark}, ABINet++ (LV) outperforms other methods consistently on all the evaluation benchmarks.

\subsection{Semi-Supervised Training}

\begin{figure}
   \begin{center}
      \includegraphics[width=0.5\textwidth]{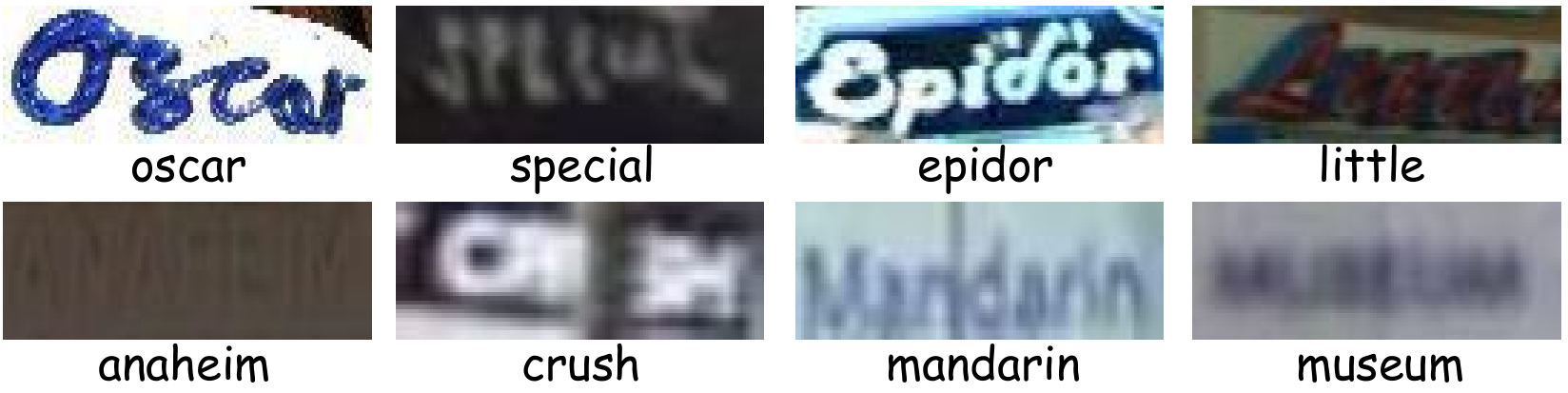}
      \caption{\textbf{Hard examples successfully recognized by ABINet++ (LV) trained with the ensemble self-training.}}
      \label{fig:semi_img}
   \end{center}
   \vspace{-2em}
\end{figure}

To further push the boundary of accurate reading, we explore a semi-supervised method which utilizes MJ and ST as the labeled datasets and Uber-Text as the unlabeled dataset. The threshold $Q$ in Section~\ref{sec:semi-supervised} is set to 0.9, and the batch size of $B_l$ and $B_u$ are 256 and 128 respectively. Experiment results in Table~\ref{tab:benchmark} show that the proposed self-training method can easily outperform the supervised version on the benchmark datasets. Besides, the ensemble self-training shows a more stable performance by improving the efficiency of data utilization. Observing the boosted results we find that hard examples with scarce fonts and blurred appearance can also be recognized frequently~(Fig.~\ref{fig:semi_img}), which suggests that exploring the semi-/unsupervised learning methods is a promising direction for scene text recognition.

\section{Experiment on Scene Text Spotting}

To well evaluate the effectiveness of ABINet++ and fairly compare with other methods, we conduct experiments based on the setup of~\cite{liu2020abcnetv2,liu2020abcnet} unless specified otherwise. The scene text benchmarks include Total-Text~\cite{ch2020total}, SCUT-CTW1500~\cite{liu2019curved}, ICDAR 2015~\cite{karatzas2015icdar}, and ReCTS~\cite{zhang2019icdar}.

\begin{table*}[htb]
   \caption{Ablation study of different components on Total-Text and SCUT-CTW1500. “OSA" is the online sampling augmentation. “SAA" is the spelling alteration augmentation. “SMN" is the sequence modeling network. “Attn" is the attention module. “HFA" is the horizontal feature aggregation. “PA" is the position attention. “PCA" is the position and content attention where the content feature is iteratively refined 3 times.}
   \label{tab:ablation-e2e}
   \newcommand{\tabincell}[2]{\begin{tabular}{@{}#1@{}}#2\end{tabular}}
   \resizebox{1.\linewidth}{!}{
   \begin{tabular}{c|cccccccc|ccc|ccc}
     \whline
     \multirow{2}*{ }  & \multicolumn{8}{c|}{Components} & \multicolumn{3}{c|}{Total-Text} &  \multicolumn{3}{c}{SCUT-CTW1500}\\
         \cline{2-15}
         & SGD & Adam & Mixed & OSA & SAA & SMN & U-Net & Attn & Hmean & Impr. & FPS & Hmean & Impr. & FPS\tablefootnote{The inference time in Table~\ref{tab:ablation-e2e},~\ref{tab:ablation-e2e-lm},~\ref{tab:sota_e2e} is estimated using an NVIDIA 3090 by averaging 5 different trials.} \\
     \hline
     (a)   & \checkmark &  &  &  &  & Trans. & w/o HFA & PA & 71.5 & - & 12.3 & 44.1 & - & - \\     
     (b)  &  & \checkmark &  &  &  & Trans. & w/o HFA & PA & 72.6 & \bf $\uparrow$ 1.1\% & 12.2 & 44.4 & \bf $\uparrow$ 0.3\% & - \\
     (c)  &  &  & \checkmark &  &  & Trans. & w/o HFA & PA & 72.9 & \bf $\uparrow$ 1.4\% & 12.2 & 46.5 & \bf $\uparrow$ 2.4\% & - \\
     \hline
     (d)  & \checkmark &  &  &  &  & Trans. & None & PA & 69.7 &  \bf $\downarrow$ 1.8\% & 12.9 & 43.7 & \bf $\downarrow$ 0.4\% & - \\ 
     (e)  & \checkmark &  &  &  &  & None & w/o HFA & PA & 70.6 &  \bf $\downarrow$ 0.9\% & 12.7 & 40.7 & \bf $\downarrow$ 3.4\% & 16.8 \\         
     (f)  & \checkmark &  &  &  &  & None & None & PA & 65.6 &  \bf $\downarrow$ 5.9\% & 13.1 & 40.5 & \bf $\downarrow$ 3.6\% & 17.5 \\
     \hline     
     (g)  &  &  & \checkmark & \checkmark &  & Trans. & w/o HFA & PA & 73.6 & \bf $\uparrow$ 2.1\% & 12.2 & 46.8 & \bf $\uparrow$ 2.7\% & - \\
     (h)  &  &  & \checkmark & \checkmark & \checkmark & Trans. & w/o HFA & PA & 74.0 & \bf $\uparrow$ 2.5\% & 12.1 & 47.7 & \bf   $\uparrow$ 3.6\% & - \\
     \hline     
     (i)  &  &  & \checkmark & \checkmark & \checkmark & Conv. & w/o HFA & PA & 76.2 & \bf $\uparrow$ 4.7\% & 12.6 & 45.5 & \bf $\uparrow$ 1.4\% & 16.5 \\
     (j)  &  &  & \checkmark & \checkmark & \checkmark & Conv. & w/o HFA & PCA & 76.1 & \bf $\uparrow$ 4.6\% & 12.4  & 51.5 & \bf $\uparrow$ 7.4\% & 16.4\\
     (k)  &  &  & \checkmark & \checkmark & \checkmark & Conv. & HFA & PA & 76.4 & \bf $\uparrow$ 4.9\% & 12.0  & 55.8 & \bf $\uparrow$ 11.7\% & 15.9\\           
     (l)  &  &  & \checkmark & \checkmark & \checkmark & Conv. & HFA & PCA & 76.3 & \bf $\uparrow$ 4.8\% & 11.8  & 59.4 & \bf $\uparrow$ 15.3\%  & 15.6\\              
     \whline
   \end{tabular}}
   \vspace{-1.em}
 \end{table*}

\begin{table}[htb]
   \caption{Ablation study of the language model on Total-Text and SCUT-CTW1500.}
   \label{tab:ablation-e2e-lm}
   \newcommand{\tabincell}[2]{\begin{tabular}{@{}#1@{}}#2\end{tabular}}
   \resizebox{1.\linewidth}{!}{
   \begin{tabular}{c|c|c|c|ccc}
     \whline
     \multirow{2}*{ } & Evaluation & \multirow{2}*{LM} & LM Fine-tuned & \multirow{2}*{Hmean} & \multirow{2}*{Impr.} & \multirow{2}*{FPS} \\
     & Dataset &  & Dataset &  \\
     \hline
     (a) & \multirow{4}{*}{Total-Text} & \ding{55} & - & 70.2 & - & 13.0 \\
     (b) & & \checkmark & SCUT-CTW1500 & 75.4 & \bf $\uparrow$ 5.2\% & - \\
     (c) & & \checkmark & ICDAR 2015 & 76.0 & \bf $\uparrow$ 5.8\% & - \\           
     (d) & & \checkmark & Total-Text & 76.3 & \bf $\uparrow$ 6.1\% & 11.8 \\ 
     \hline
     (e) & & \ding{55} & - & 53.5 & - & 17.5 \\
     (f) & SCUT- & \checkmark & Total-Text & 57.2 & \bf $\uparrow$ 3.7\%  & - \\
     (g) & CTW1500 & \checkmark & ICDAR 2015 & 57.0 & \bf $\uparrow$ 3.5\% & - \\
     (h) & & \checkmark & SCUT-CTW1500 & 59.4 & \bf $\uparrow$ 5.9\% & 15.6 \\              
     \whline     
   \end{tabular}}
 \end{table}

\subsection{Implementation Details}

The settings of the detection network follow previous works~\cite{liu2020abcnetv2,liu2020abcnet}. For English recognition, models are first pre-trained using SynthText150K~\cite{liu2020abcnet}, ICDAR MLT~\cite{nayef2017icdar2017}, and the training set of each dataset annotated at word-level, and then fine-tuned on each training set. There are 96 character classes for English recognition, including digits, case-sensitive letters and 34 ASCII punctuation marks. For Chinese recognition (ReCTS), models are pre-trained using SynChinese130K~\cite{liu2020abcnet}, ReCTS~\cite{zhang2019icdar}, LSVT~\cite{sun2019icdar} and ArT~\cite{chng2019icdar2019}. There are 5,461 character classes in total for Chinese recognition. We set the maximum length of output text to 25 for regular-length datasets, and otherwise 100 for long text dataset (\ie, CTW1500). Correspondingly, the pooled sizes of BezierAlign are set to $8 \times 32$ and $8 \times 128$, respectively.

The SMN is implemented by a 4-layer convolution network. The number of Transformer layers inside the U-Net in Table~\ref{tab:unet} is empirically set to 4. $p_r, p_i, p_d, p_u$ in Section~\ref{sec:text-aug} are set to $0.2, 0.05, 0.05, 0.7$, respectively.

Generally, Stochastic Gradient Descent (SGD) generalizes better on many computer vision tasks~\cite{zhou2020towards, he2017mask}, while Adam outperforms SGD on attention models (\eg, Transformer\cite{vaswani2017attention}, BERT\cite{devlin2018bert}) due to its faster convergence speed~\cite{zhang2019adam}. Therefore, we adopt an optimizer-mixed method to train our text spotter. Specifically, the VM is trained using SGD, while the LM and the fusion model are trained using Adam. At the pre-training stage, the initial learning rates for SGD and Adam are 1e-2 and 1e-4 respectively, which decays to a tenth at 400K$^{\rm th}$ and 500K$^{\rm th}$ iterations. Training is stopped at 540K$^{\rm th}$ iteration. At the fine-tuning stage, the learning rates start from 1e-3 and 1e-5, respectively. Models are trained using 4 NVIDIA 3090 GPUs. The batch size is set to 8, and the batch size $B_l$ for the LM is 384.

\subsection{Datasets}

\textbf{Total-Text}~\cite{ch2020total} is an arbitrarily-shaped dataset which consists of 1,255 training images and 300 test images. \textbf{SCUT-CTW1500}~\cite{liu2019curved} is an arbitrarily-shaped dataset containing long text images. There are 1,000 images for training and 500 images for evaluation. \textbf{ICDAR 2015}~\cite{karatzas2015icdar} is a multi-oriented dataset including 1,000 training images and 500 testing images. \textbf{ICDAR 2019 ReCTS}~\cite{zhang2019icdar} is a Chinese dataset, where 20K images are used for training and 5K images are used for evaluation.

Besides, several datasets are introduced for training only, including a synthetic dataset \textbf{SynthText150K}~\cite{liu2020abcnetv2,liu2020abcnet} that has 150K images, a synthetic dataset \textbf{SynChinese130K}~\cite{liu2020abcnetv2} consisting of about 130K images, a multi-lingual dataset \textbf{ICDAR 2017 MLT}~\cite{nayef2017icdar2017} with 7.2K training images, an arbitrarily-shaped dataset \textbf{ICDAR 2019 ArT}~\cite{chng2019icdar2019} containing 5,603 training images, and a large-scale dataset \textbf{ICDAR 2019 LSVT}~\cite{sun2019icdar} where 30K labeled images are used. The above datasets are detailed in the supplementary material.

\subsection{Ablation Study}
\label{sec:exp_e2e_ablation}

Ablation studies are conducted on Total-Text and SCUT-CTW1500, which are responsible for evaluating performance on regular-length text and long text images. Experiment statistics are recorded in Table~\ref{tab:ablation-e2e}. The testing scales for Total-Text and SCUT-CTW1500 are 1000 and 800 on the short side.

Firstly, three kinds of optimizers are compared in Table~\ref{tab:ablation-e2e} (a-c), \ie, SGD, Adam, and the Mixed that applies SGD for detection and Adam for recognition. We find that the optimizer Mixed method shows better performance than using only single ones, as 1) we observe the training loss of the LM using SGD decreases more slowly than Adam. 2) Optimizing the detection module using Adam causes a worse generalization performance. There are about $0.6\%$ and $2.3\%$ gaps in text detection in terms of Hmean on Total-Text and SCUT-CTW1500, respectively.

Then we evaluate several components of ABINet++ in Table~\ref{tab:ablation-e2e} (d-f). By removing U-Net from the attention module, we observe a drop of $1.8\%$ on Total-Text and $0.4\%$ on SCUT-CTW1500, respectively. In a similar vein, without the use of SMN, Hmean decreases by $0.9\%$ and $3.4\%$. This prompts us for a regular-length dataset, U-Net acts as an important role in learning discriminative features. Nevertheless, for a long text dataset, the SMN implemented by a Transformer\footnote{As the $8 \times 128$ pooled size for SCUT-CTW1500 dataset is too big to train these models, we simply use $2 \times 128$ for the corresponding models.} appears to be more effective compared with U-Net. This phenomenon can be partly caused by the Transformer which is able to aggregate features along character order via non-local mechanism, and we further discuss it in Section~\ref{sec:exp_e2e_long_text}. Besides, discarding both parts simultaneously brings a substantial impact on Hmean. From the observation we can conclude that despite deep and multi-scale features obtained by shared FPN, the feature maps for the recognition branch are still badly in need of recognition-oriented abstraction, \ie, a well-designed structure after the RoI extractor, which greatly deviates from the detection branch, is highly necessary.

Based on the optimizer-mixed method we evaluate the text augmentation methods of LM proposed in Section~\ref{sec:text-aug}. As can be seen in Table~\ref{tab:ablation-e2e}~(g-h), the OSA and SAA can further consistently improve the performance by $0.7\%$ and $1.1\%$ on Total-Text, and $0.3\%$ and $1.2\%$ on SCUT-CTW1500, which demonstrates their effectiveness to reinforce the learning of language models.

Further, we replace the Transformer units with the convolution layers as SMN in Table~\ref{tab:ablation-e2e}~(i), not only because of the obvious benefit on regular-length datasets (\eg, $2.2\%$ improvement on Total-Text), but also as the convolution can efficiently deal with long sequences, which the Transformer is incapable of due to quadratic time complexity and high memory usage. Based on the above settings, the horizontal feature aggregation (HFA) and position and content attention (PCA) are evaluated separately and jointly. The results presented in Table~\ref{tab:ablation-e2e}~(j-l) prove the proposed HFA and PCA can substantially increase the Hmean performance of ABINet++ on SCUT-CTW1500. From the visualization of attention maps in Fig.~\ref{fig:attention_map} and Fig.~\ref{fig:e2e-img} we can see, HFA and PCA can effectively alleviate the multi-activation phenomenon. Meanwhile, both the HFA and PCA only introduce acceptable computation overhead. On the other hand, we observe few side-effects on regular-length datasets in terms of accuracy. Note that the iteration number for PCA in Table~\ref{tab:ablation-e2e} is set to 3, and more iterations might further bring a slight improvement.

Finally, the impact of the LM in an end-to-end text spotter is discussed. Compared with the methods using pure visual cures in Table~\ref{tab:ablation-e2e-lm}~(a;e), the models equipped with the proposed BCN in Table~\ref{tab:ablation-e2e-lm}~(d;h) can significantly improve Hmean by $6.1\%$ and $5.9\%$ on Total-Text and SCUT-CTW1500, respectively, at the expense of a slight decrease in speed. We also investigate out-of-domain performance of the LMs. As seen in Table~\ref{tab:ablation-e2e-lm}~(b-d;f-h), the LMs fine-tuned on out-of-domain data can still obtain satisfactory performance compared with in-domain data, which indicates our independent LMs have impressive generalization ability. We also observe that a larger gap between the source domain and target domain brings greater challenges in domain transfer. For example, SCUT-CTW1500 with long text is quite different than Total-Text and ICDAR 2015, therefore fine-tuning LM on ICDAR 2015 achieves better Hmean than on SCUT-CTW1500 when evaluated on Total-Text.

\subsection{Long Text Recognition}
\label{sec:exp_e2e_long_text}

\begin{table}[!t]
   \centering
   \caption{Analysis in long text recognition. “Size" is the pooled size of BezierAlign. “Stride" is the total downsampling stride of U-Net (refer to Table~\ref{tab:unet}). “Aggr." is the aggregation method used in U-Net. $\#n$ is the iteration number in PCA.}
   \label{tab:long_text_recognition}
   \small
   \resizebox{1.\linewidth}{!}{
   \begin{tabular}{c|cccc|c|c}
      \whline
       & Size & Stride & Aggr. & Attn & Hmean & Impr. \\
      \hline
      (a) & $(8, 128)$ & $(8, 16)$ & - & PA & 30.1 & - \\
      \hline
      (b) & $(8, 32)$ & $(8, 16)$ & - & PA & 35.4 & $\uparrow$ 5.3\% \\
      (c) & $(8, 128)$ & $(8, 128)$ & - & PA & 36.1 & $\uparrow$ 6.0\% \\
      (d) & $(8, 128)$ & $(1, 128)$ & - & PA & 36.3 & $\uparrow$ 6.2\% \\
      (e) & $(8, 128)$ & $(8, 1)$ & - & PA & 24.2 & $\downarrow$ 5.9\% \\
      (f) & $(8, 128)$ & $(1, 1)$ & - & PA & 24.2 & $\downarrow$ 5.9\% \\
      \hline
      (g) & $(8, 128)$ & $(8, 16)$ & Mean & PA & 40.8 & $\uparrow$ 10.7\% \\
      (h) & $(8, 128)$ & $(8, 16)$ & Conv-1 & PA & 42.4 & $\uparrow$ 12.3\% \\ 
      (i) & $(8, 128)$ & $(8, 16)$ & Trans-1 & PA & 43.1 & $\uparrow$ 13.0\% \\
      (j) & $(8, 128)$ & $(8, 16)$ & Trans-4 & PA & 44.0 & $\uparrow$ 13.9\% \\           
      \hline
      (k) & $(8, 128)$ & $(8, 16)$ & - & PCA\#$1$ & 37.0 & $\uparrow$ 6.9\% \\   
      (l) & $(8, 128)$ & $(8, 16)$ & - & PCA\#$2$ & 42.0 & $\uparrow$ 11.9\% \\  
      (m) & $(8, 128)$ & $(8, 16)$ & - & PCA\#$3$ & 43.2 & $\uparrow$ 12.1\% \\  
      (n) & $(8, 128)$ & $(8, 16)$ & - & PCA\#$4$ & 43.3 & $\uparrow$ 12.2\% \\  
      \hline
      (o) & $(8, 128)$ & $(8, 16)$ & Trans-4 & PCA\#$3$ & 47.3 & $\uparrow$ 17.2\% \\                                
      \whline
   \end{tabular}}
 \end{table}

To gain a deep insight into long text recognition, we experimentally explore potential causes that would degrade the performance. Experiments are directly conducted on SCUT-CTW1500 without pre-training on other datasets. The experiment results are recorded in Table~\ref{tab:long_text_recognition}, where all the models apply convolution layers as SMN.

Based on the baseline model in Table~\ref{tab:long_text_recognition} (a), we first replace the pooled size of BezierAlign with a smaller size $8 \times 32$ like on regular-length datasets. We find the Hmean boosts by $5.3\%$ immediately, as seen in Table~\ref{tab:long_text_recognition} (b). However, this is unreasonable for long text recognition with a character number greater than $32$, since each location in the feature maps is unlikely to encode more than two characters. Observing the difference that the sizes of the finest feature maps inside U-Net are $1 \times 2$ and $1 \times 8$ for pooled size $8 \times 32$  and $8 \times 128$ respectively, we can infer an assumption that aggregating information along character direction into a compact feature representation is significant for the position attention. As the downsampling operation in U-Net can be viewed as a feature aggregation method, by evaluating the impact of different U-Net strides we can further validate this assumption. As reported in Table~\ref{tab:long_text_recognition} (c-f), increasing the horizontal stride to $128$ can bring over $6$ points improvement, while decreasing the horizontal stride harms the accuracy significantly. Besides, compared with the vertical stride, the horizontal stride along character direction greatly influences the Hmean.

Based on this observation, we further explore some aggregation methods for horizontal features which are more general than manipulating strides (Table~\ref{tab:long_text_recognition} (g-j)). For a U-Net with a stride of $8 \times 16$, we embed operation layers after the finest features ($1 \times 8$) to aggregate information. Interestingly, we find that a simple mean operation can improve Hmean by $10.7\%$ compared with the baseline model. The performance could be further boosted by using a convolution layer with kernel size $1 \times 8$, or a single Transformer layer. Considering the efficiency and scalability for arbitrary length text, we finally embed a Transformer with 4 layers into the U-Net.

In the meanwhile, we show that by introducing the content feature, the proposed PCA can improve the discrimination of the query vector and thus bring a fundamental change in long text recognition, as reported in Table~\ref{tab:long_text_recognition} (k-n). Even if a coarse context vector is directly used, the performance can be improved by $6.9\%$ in terms of Hmean. With the aid of the iteration strategy, the content vector is refined gradually and Hmean can be improved by up to $12.2\%$. Finally, by combing HFA and PCA, the upgraded model increases Hmean from $30.1\%$ to $47.3\%$, achieving a competitive performance even without pre-training.

\begin{table*}[!t]
   \caption{Performance comparison with other methods for end-to-end scene text spotting. ``None'' indicates lexicon-free. ``Full'' represents the lexicon that includes all the words in the test set. ``S'', ``W'', and ``G'' mean recognition with ``Strong'', ``Weak'', and ``Generic'' lexicon, respectively. $^*$ denotes requiring character-level annotations. ``MS'' is multi-scale result.}
   \label{tab:sota_e2e}
   \centering
   \newcommand{\tabincell}[2]{\begin{tabular}{@{}#1@{}}#2\end{tabular}}
   \small
   \resizebox{1.\linewidth}{!}{
   \begin{tabular}{ r |c|cc|cc|cccc|c|c}
      \whline
      \multirow{2}*{Methods} & \multirow{2}*{Year} & \multicolumn{2}{c|}{Total-Text} & \multicolumn{2}{c|}{SCUT-CTW1500} &
      \multicolumn{4}{c|}{ICDAR 2015 End-to-End}  &
      \multicolumn{1}{c|}{ReCTS}  & \multirow{2}*{FPS}\\
      \cline{3-11}
      &   & None & Full & None & Full  & S & W & G & None &1-NED \\
     \hline
     FOTS~Liu~\etal~\cite{liu2018fots} & 2018&-&- &21.1&39.7  &83.6&79.1&65.3&- &50.8  &-\\     
     Qin~\etal~\cite{qin2019towards} & 2019 &67.8&- &-&-  &-&-&-&-  &- &4.8\\
     CharNet~Xing~\etal$^*$~\cite{xing2019convolutional} & 2019 &-&-  &-&- &80.1&74.5&62.2&- &-  &1.2\\
     TextDragon~Feng~\etal~\cite{feng2019textdragon} & 2019&48.8&74.8 &39.7&72.4  &82.5&78.3&65.2&- &- &2.6\\
     Mask TextSpotter'19~Liao~\etal~\cite{liao2019mask} & 2019 &65.3&77.4  &-&- &83.0&77.7&73.5&- &67.8  &2.0\\     
     ABCNet~Liu~\etal~\cite{liu2020abcnet} & 2020&64.2&75.7 &45.2&74.1  &-&-&-&- &- & 17.9\\
     Boundary~Wang~\etal\cite{wang2020all} & 2020&-&- &-&-  &79.7&75.2&64.1&- &- &-\\
     Craft~Baek~\etal$^*$~\cite{baek2020character} & 2020&78.7&- &-&-  &83.1&82.1&74.9&- &- &5.4\\
     Mask TextSpotter v3~Liao~\etal~\cite{liao2020mask} & 2020&71.2&78.4 &-&-  &83.3&78.1&74.2&- &- &2.5\\
     Feng~\etal~\cite{feng2021residual} & 2021&55.8&79.2 &42.2&74.9 &\bf{87.3}&\bf{83.1}&69.5&- &-&7.2\\
     PGNet~Wang~\etal~\cite{wang2021pgnet} & 2021& 63.1&- &-&- &83.3&78.3&63.5&- &- & \bf{35.5}\\
     MANGO~Qiao~\etal~\cite{qiao2021mango} & 2021& 72.9&83.6 &58.9&78.7 &85.4&80.1&73.9&- &- &4.3\\ 
     PAN++~Wang~\etal~\cite{wang2021pan++} & 2021& 68.6&78.6 &-&- &82.7&78.2&69.2&68.0 &- &21.1 \\
     ABCNet v2~Liu~\etal~\cite{liu2020abcnetv2} & 2021 &70.4&78.1  &57.5&77.2 &82.7&78.5&73.0&-  &62.7 &10\\
     \hline
     ABINet++ & &77.6&84.5 &60.2&80.3 &84.1&80.4&75.4&73.3 &76.5 &10.6\\
     ABINet++~MS & &\bf{79.4}&\bf{85.4} &\bf{61.5}&\bf{81.2} &86.1&81.9&\bf{77.8}&\bf{74.7} &\bf{77.4} & -\\
     \whline
   \end{tabular}}
   \vspace{-2em}
 \end{table*}
 
 \begin{figure}
   \begin{center}
      \includegraphics[width=0.5\textwidth]{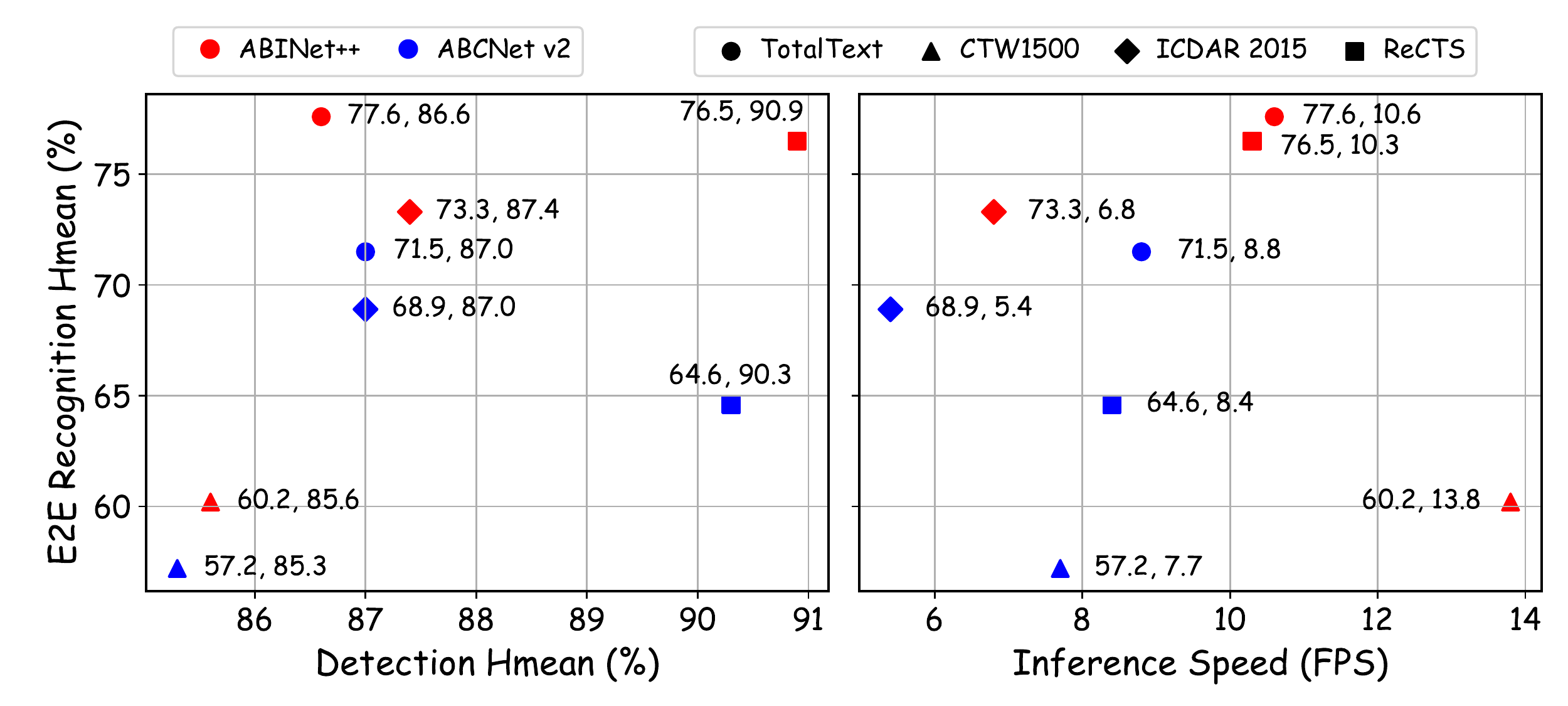}
      \caption{\textbf{Comparison between ABCNet v2 and ABINet++.}}
      \label{fig:comparison}
   \end{center}
   \vspace{-1.5em}
 \end{figure}

\subsection{Comparisons with State-of-the-Arts}

To comprehensively show the superiority of ABINet++ in scene text spotting, we compare our method with recently state-of-the-art methods. Based on the conclusion of the above discussion, we further optimize ABINet++ from the aspects of inference thresholds, testing scales and repeating the BiFPN block by 2 times. The evaluation datasets include Total-Text, SCUT-CTW1500, ICDAR 2015 and ReCTS datasets. All the evaluation metrics follow previous methods~\cite{liu2020abcnetv2,wang2021pan++,liao2019mask}. As our method focuses on the recognition issues in text spotting, we omit the detection accuracy and only compare the end-to-end recognition performance. Comparison results are recorded in Table~\ref{tab:sota_e2e}.

\textbf{Total-Text}. ABINet++ achieves a Hmean of $77.6\%$, which significantly exceeds the recent state-of-the-art methods that are without the need of character-level annotations by at least $4.7\%$. Meanwhile, our method maintains real-time speed and runs at 10.6 FPS using an NVIDIA 3090 GPU. These results prove that, compared with the CTC-based~\cite{liu2020abcnet, wang2021pgnet}, attention-based~\cite{liu2020abcnetv2,wang2021pan++} and classification-based~\cite{qiao2021mango} text spotters, ABINet++ not only presents excellent recognition accuracy, but also obtains comparable inference speed on arbitrarily-shaped text spotting.

\textbf{SCUT-CTW1500}. Our method surpasses the state-of-the-art methods by at least $1.3\%$ and $1.6\%$ in “None” and “Full” metrics, showing that ABINet++ can effectively deal with long text recognition.

\textbf{ICDAR 2015}. ABINet++ achieves the best performance on ICDAR 2015 using the “None” and “Generic” lexicon, which demonstrates the good generalization of ABINet++ in different challenging environments.

\textbf{ReCTS}. Even directly applying the settings on Total-Text to ReCTS, ABINet++ can still achieve impressive performance, significantly boosting the 1-NED metric from $67.8\%$ (Mask TextSpotter~\cite{liao2019mask}) to $76.5\%$. This indicates ABINet++ has a huge potential in recognizing Chinese text.

In addition, as the reported performances from original literature diverge from training datasets, training strategies (\eg, fine-tuning, jointly training), image scales both in training and inference, \etc., it is unrealistic to analyze the significant superiority of ABINet++ over different methods. Therefore, we compare our method with ABCNet v2 under the same environments for a strictly fair comparison. Both the reproduced ABCNet v2 and ABINet++ are implemented using the same repository\footnote{https://github.com/aim-uofa/AdelaiDet}, where we share the same experiment configurations and Bezier detection module. As seen in Fig.~\ref{fig:comparison}, despite having similar detection accuracy with ABCNet v2, our method consistently and significantly improves the performance of ABCNet v2 in terms of Hmean on four datasets. Specifically, the Hmeans are improved by $6.1\%$, $3.0\%$, $4.4\%$ and $11.9\%$ on Total-Text, SCUT-CTW1500, ICDAR 2015 and ReCTS, respectively. Some qualitative examples of ABINet++ are presented in Fig.~\ref{fig:e2e-img}. Note that the CoordConv proved to be effective for end-to-end accuracy~\cite{liu2020abcnetv2} is abandoned in our work due to the limitation of computation resources. From the point of inference efficiency, ABINet++ shows faster speed compared with ABCNet v2 due to the parallel recognition structure and without the use of CoordConv. In summary, the statistics indicate a powerful recognition model incorporating effective language modeling is significant for end-to-end scene text spotting.


\section{Discussion}

Despite benefitting from linguistic knowledge, current LMs in scene text recognition and spotting, including ABINet++, basically learn the characters relationship (\ie, the spelling conventions in linguistics) rather than the semantics between tokens/words. However, high-level semantic information is also useful for recognition, especially in long text recognition. For example, for an image containing a text string ``I love cat and he loves dog'', if the ``cat'' is mistakenly recognized as ``hat'', it is unlike for a character-level LM to correct the ``hat'' to the ``cat'' since the ``hat'' is spelled correctly. However, a token-level LM can learn semantic context and therefore infer the correct written form. There is still a long way to go for language modeling in text recognition.

Another issue we should point out is that currently ABINet++ converges more slowly than other spotters, even though we use the optimizer-mixed training strategy. For example, the iteration number in the pre-training stage is more than 2 times of ABCNet v2. Besides, there exists an optimization gap between the VM and LM. Therefore, how to effectively optimize a cross-modal model to further improve its performance remains a challenge.

\begin{figure*}
   \begin{center}
      \includegraphics[width=1.0\textwidth]{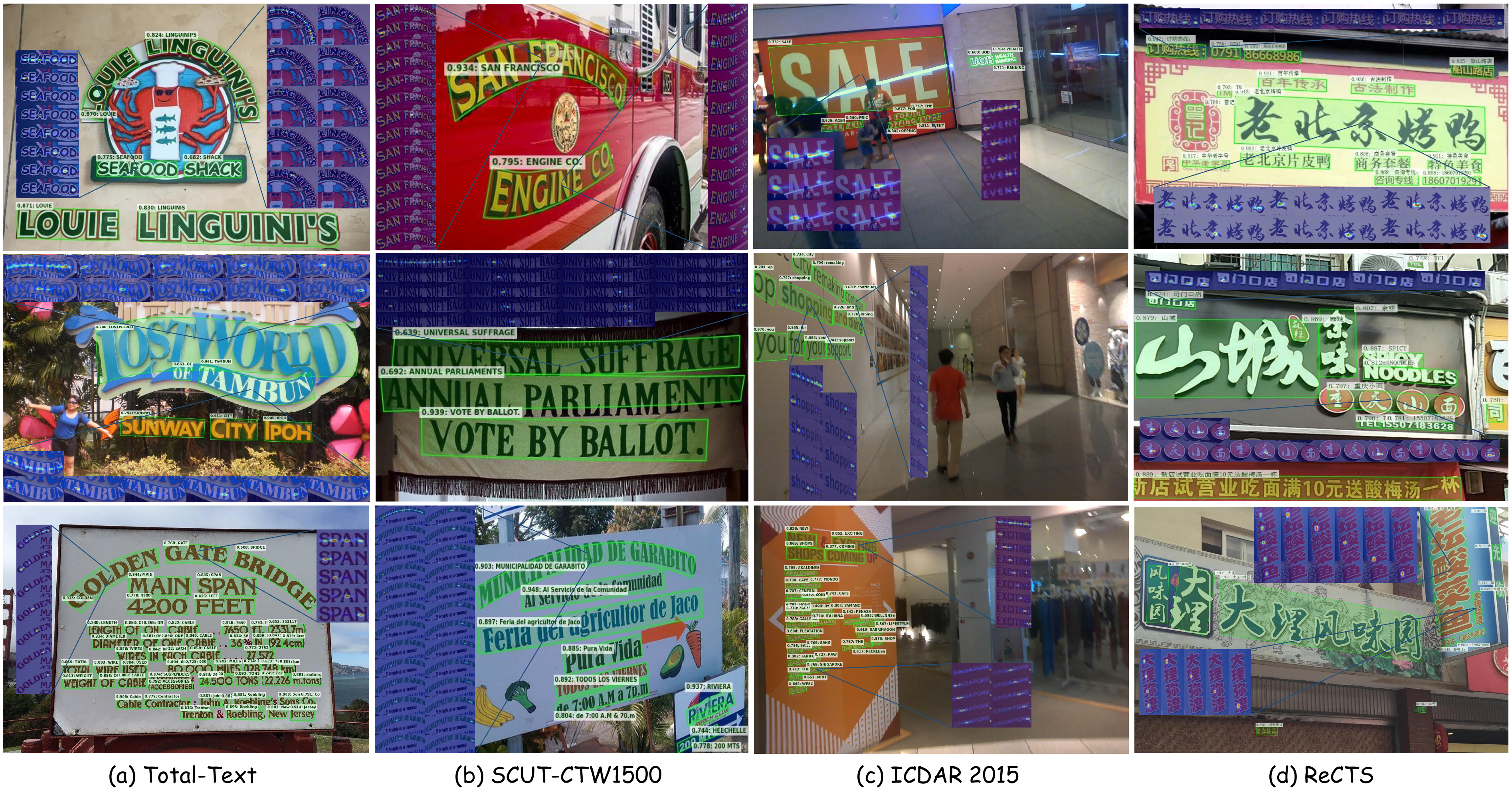}
      \caption{\textbf{Qualitative text spotting results of ABINet++ on various datasets.} The attention maps are also restored for visualization.}
      \label{fig:e2e-img}
   \end{center}
   \vspace{-1em}
 \end{figure*}

For scene text recognition, the common errors include unusual font, text in weak semantics, extremely long text images, \etc. For scene text spotting, errors might additionally come from inaccurate detection results, such as vertical text images, and text patches whose top and bottom boundaries are wrongly determined, \etc. The failure cases could be found in the supplementary material. Note that for both above tasks, the accuracy of the VM is still an important factor for overall performance. We also observe that in the text spotting task, the attention maps learned by the VM tend to be worse than the attention maps in the text recognition task by comparing the visualizations in Fig.~\ref{fig:attention_map}, Fig.~\ref{fig:long-text} and Fig.~\ref{fig:e2e-img}, due to the more complicated environments in scene text spotting.

\section{Conclusion}

In this paper, we propose ABINet++ which explores effective approaches for utilizing linguistic knowledge in scene text spotting. The ABINet++ is 1) autonomous which improves the ability of the language model by enforcing learning explicitly, 2) bidirectional which learns text representation by jointly conditioning on character context at both sides, and 3) iterative which corrects the prediction progressively to alleviate the impact of noise input. Besides, we propose horizontal feature aggregation and position and content attention to effectively improve the performance on long text images. We also introduce the methods of pre-training the language model from external text datasets and sampling misspelled text online to reinforce the learning of the language model. 

Extensive experiments are conducted, including on regular and irregular word-box datasets, arbitrarily-shaped and multi-oriented datasets, regular-length text and long text datasets, English and Chinese datasets to demonstrate the superiority of ABINet++. Besides, different from the previous text spotters which mainly focus on text detection, we claim that a strong recognizer is also significant to the performance of an end-to-end text spotter.

\section*{Acknowledgments}

This work is supported by the National Natural Science Foundation of China (62102384, 62121002, 62022076, U1936210, 62222212).

\ifCLASSOPTIONcaptionsoff
  \newpage
\fi

{
\bibliographystyle{IEEEtran}
\bibliography{IEEEabrv}
}

%




\end{document}


\title{Supplementary Material}

\maketitle

\section{Analysis of state-of-the-art method}

\begin{table}[htp]
   \begin{center}
   \caption{Ratios of failure cases (\%) of SOTA method~\cite{yu2020towards}. Strong/Weak Semantics is estimated by classifying text inside/outside Oxford Dictionaries. $ed$ is the edit distance.}
   \label{tab:sota-bad-cases}
   \resizebox{1.0\linewidth}{!}{
   \begin{tabular}{|c|c|c|c|c|c|}
      \hline
      \multicolumn{3}{|c|}{Strong Semantics}  & \multicolumn{2}{|c|}{Weak Semantics} & \multirow{1}{*}{Illegible \&}  \\
      \cline{1-5}
      $ed$=1 & $ed$=2 & $ed \ge 3$ & alphabet & digit & Wrong Label  \\
      \hline
      23.2 & 11.0 & 11.6 & 26.9 & 3.6 &  23.7 \\
      \hline
   \end{tabular}}
   \end{center}
\end{table}

We analyze the failure cases of the current state-of-the-art (SOTA) method SRN~\cite{yu2020towards}, as the statistical data shown in Table~\ref{tab:sota-bad-cases}. Among the failure cases, $45.8\%$ text is with strong semantics, which should be successfully reasoned by linguistic rules. Specifically, $23.2\%$ text is wrongly recognized by SRN where the edit distance between predicted text and ground truth is only 1. This denotes that there is still a big room for enhancing SOTA methods in language modeling, and therefore we propose the ABINet++ aiming to improve the ability of language modeling.

\section{Comparison of different language models}

As shown in Table~\ref{tab:comparison_lm}, we compare the BCN in ABINet++ with other mainstream language models in multiple aspects. In the field of scene text recognition, SRN~\cite{yu2020towards} proposes a unidirectional GPT-like language model, which adopts causal masks and combines two unidirectional models. Taking left-to-right directional decoding as an example, the prediction of each time step is based on the predictions of the previous time steps and the input of the current time step. For example, taking a sequence of \text{{\tt{[S]}},A,B,C} as input, \text{A,B,C,{\tt{[S]}}} are predicted one by one, where {\tt{[S]}} is a special token representing the start position or end position. This kind of directional-dependant language model cannot learn bidirectional representation for language features and thus cannot make full use of contextual information.

The masking strategy both in BERT~\cite{devlin2018bert} and PIMNet~\cite{qiao2021pimnet} is masked language model (MLM), which can realize bidirectional feature representation. BERT~\cite{devlin2018bert} can be applied to text recognition, in which the mispredicted characters are replaced with {\tt{[MASK]}} and then corrected by BERT prediction. However, in this case, BERT~\cite{devlin2018bert} needs $n$ times network passes for character decoding, where $n$ is the text sequence length. This method significantly consumes computing resources and increases inference time. PIMNet~\cite{qiao2021pimnet} further extends it by gradually generating characters from a sequence of {\tt{[MASK]}} within $k\enspace(1 \le k \le n$) times iterations. During both the training and inference stage, PIMNet~\cite{qiao2021pimnet} adopts an easy-first strategy that gets the most confident predictions in each iteration. 

The above methods are all based on the Transformer encoder, while the structures of DisCo~\cite{kasai2020nonautoregressive} and BCN are the variants of the Transformer decoder. Both of them are based on another kind of masking strategy, called attention-masking model (AMM), and can model bidirectional feature representation. However, as DisCo~\cite{kasai2020nonautoregressive} is a machine translation model which cannot obtain a coarse prediction initially as in scene text recognition (\ie, we can obtain it using a vision model), thus the design of DisCo~\cite{kasai2020nonautoregressive} is quite different from our spelling correction model BCN: firstly, during the training phase, DisCo~\cite{kasai2020nonautoregressive} adopts Randomly Sampled Masks in self-attention, which predicts tokens using a subset of the other reference tokens that are sampled randomly. Differently, BCN uses a deterministic process by presetting a diagonal mask simulating the cloze test, where the attention-masking is performed in the cross-attention layer rather than the self-attention as in DisCo. Secondly, to avoid information leakage, DisCo~\cite{kasai2020nonautoregressive} carefully designs its structure by decontextualizing key and value vectors, while BCN gives up self-attention layers to achieve the same goal. Thirdly, the training and inference stages of BCN perform in the same way and it does not require multiple network passes for character decoding due to the deterministic diagonal masking operation. However, the inference stage in DisCo~\cite{kasai2020nonautoregressive} is different from the training stage thus an additional Parallel Easy-First algorithm is needed. As a result, the network passes of Disco in the inference stage need $t$ times, where $t$ is the max number of iteration preset initially.

\begin{table*}[htp]
   \centering
   \caption{Comparison of different language models.}
   \label{tab:comparison_lm}
   \resizebox{1.\linewidth}{!}{
   \begin{tabular}{cccccc}
      \whline\noalign{\smallskip}
      Language Model & SRN\cite{yu2020towards} & BERT\cite{devlin2018bert} & PIMNet\cite{qiao2021pimnet} & DisCo\cite{kasai2020nonautoregressive} & BCN \\
      \noalign{\smallskip}\hline\noalign{\smallskip}
      \makecell{Diagram\\ \\(Training)}
      & \begin{minipage}[b]{0.25\columnwidth}
            \centering
            \raisebox{-.5\height}{\includegraphics[width=\linewidth]{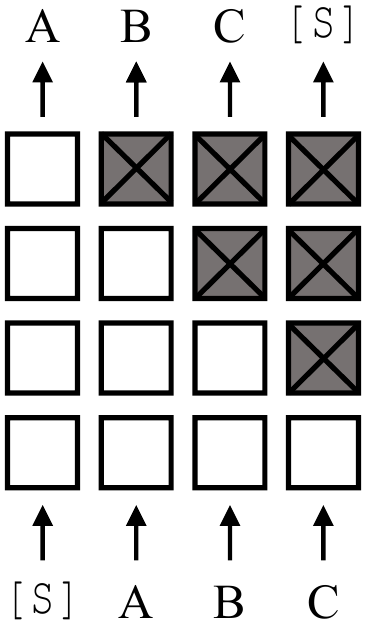}}
         \end{minipage}
      & \begin{minipage}[b]{0.25\columnwidth}
            \centering
            \raisebox{-.5\height}{\includegraphics[width=\linewidth]{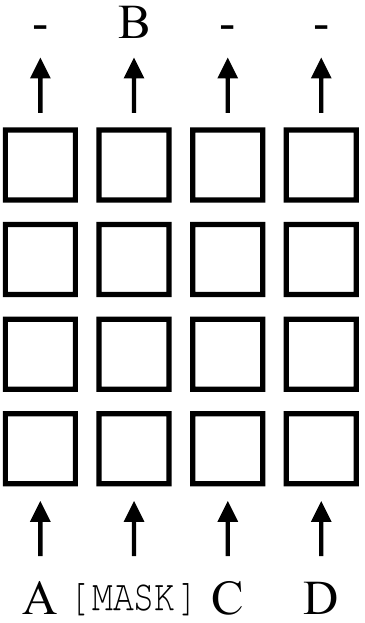}}
         \end{minipage}
      &  \begin{minipage}[b]{0.35\columnwidth}
            \centering
            \raisebox{-.5\height}{\includegraphics[width=\linewidth]{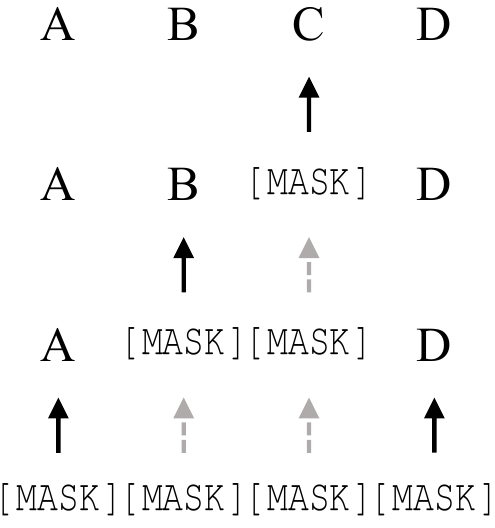}}
         \end{minipage}
      & \begin{minipage}[b]{0.25\columnwidth}
            \centering
            \raisebox{-.5\height}{\includegraphics[width=\linewidth]{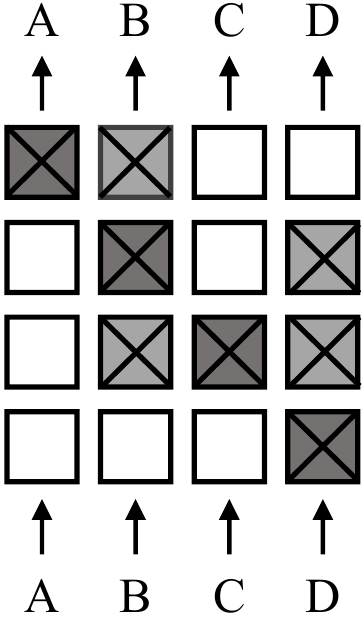}}
         \end{minipage}
      & \begin{minipage}[b]{0.25\columnwidth}
            \centering
            \raisebox{-.5\height}{\includegraphics[width=\linewidth]{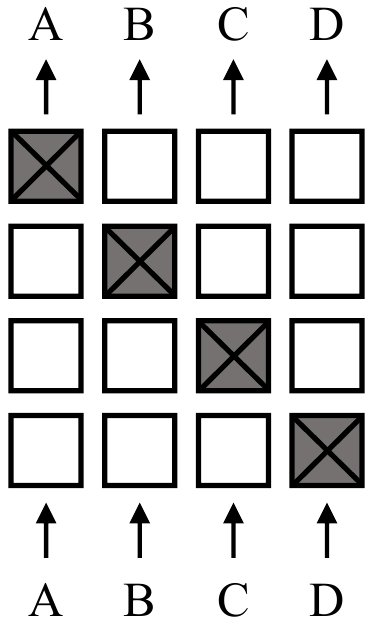}}
         \end{minipage} \\
      \noalign{\smallskip}\hline\noalign{\smallskip}
      Structure & Transformer Encoder & Transformer Encoder & Transformer Encoder & Transformer Decoder & Transformer Decoder \\      
      \noalign{\smallskip}\hline\noalign{\smallskip} 
      Feature Representation & Unidirectional & Bidirectional & Bidirectional & Bidirectional &  Bidirectional \\
      \noalign{\smallskip}\hline\noalign{\smallskip}
      Masking Strategy & GPT-like & MLM & MLM & AMM in Self-Attention & AMM in Cross-Attention \\
      \noalign{\smallskip}\hline\noalign{\smallskip} 
      Mask Type & 
      Casual Masks &
      {\tt{[MASK]}} Tokens &
      {\tt{[MASK]}} Tokens &
      Randomly Sampled Masks &
      Diagonal Masks \\
      \noalign{\smallskip}\hline\noalign{\smallskip} 
      Training & 
      $\begin{aligned}
         \text{Input}:&\enspace \text{{\tt{[S]}},A,B,C} \\
         T_0:&\enspace P(\text{A} \vert \text{{\tt{[S]}}}) \\ 
         T_1:&\enspace P(\text{B} \vert \text{{\tt{[S]}},A}) \\
         T_2:&\enspace P(\text{C} \vert \text{{\tt{[S]}},A,B}) \\
         T_3:&\enspace P(\text{{\tt{[S]}}} \vert \text{{\tt{[S]}},A,B,C})
      \end{aligned}$ &
      $\begin{aligned}
         \text{Input}:&\enspace \text{A,B,C,D} \\
         T_0:&\enspace P(\text{A} | \text{{\tt{[M]}},B,C,D}) \\ 
         T_1:&\enspace P(\text{B} | \text{A,{\tt{[M]}},C,D}) \\
         T_2:&\enspace P(\text{C} | \text{A,B,{\tt{[M]}},D}) \\
         T_3:&\enspace P(\text{D} | \text{A,B,C,{\tt{[M]}}})
      \end{aligned}$ &
      $\begin{aligned}
         \text{Input}:&\enspace \text{{\tt{[M]}},{\tt{[M]}},{\tt{[M]}},{\tt{[M]}}} \\
         Iter_1:&\enspace P(T_{i}|\text{{\tt{[M]}},{\tt{[M]}},{\tt{[M]}},{\tt{[M]}}}) \\ 
         Iter_2:&\enspace P(T_{i}|\text{A,{\tt{[M]}},{\tt{[M]}},D}) \\
         Iter_3:&\enspace P(T_{i}|\text{A,B,{\tt{[M]}},D}) \\
         \\
      \end{aligned}$ &
      $\begin{aligned} 
         \text{Input}:&\enspace \text{A,B,C,D} \\
         T_0:&\enspace P(\text{A} | \text{C,D}) \\ 
         T_1:&\enspace P(\text{B} | \text{A,C}) \\
         T_2:&\enspace P(\text{C} | \text{A}) \\
         T_3:&\enspace P(\text{D} | \text{A,B,C})&
      \end{aligned}$ &
      $\begin{aligned}
         \text{Input}:&\enspace \text{A,B,C,D} \\
         T_0:&\enspace P(\text{A} | \text{B,C,D}) \\ 
         T_1:&\enspace P(\text{B} | \text{A,C,D}) \\
         T_2:&\enspace P(\text{C} | \text{A,B,D}) \\
         T_3:&\enspace P(\text{D} | \text{A,B,C}) \\
      \end{aligned}$ \\
      \noalign{\smallskip}\hline\noalign{\smallskip}
      Inference & 
      \makecell{Identify with Training}&
      \makecell{Identify with Training}&
      \makecell{Identify with Training \\(Easy-First)}&
      \makecell{Different from Training \\(Easy-First)}&
      \makecell{Identify with Training} \\ 
      \noalign{\smallskip}\hline\noalign{\smallskip}
      \makecell{Network Passes \\ in Inference} & $1$ & $n$ (Sequence Length)  & $k\enspace(1 \le k \le n$) &
      $t$ (Initial Setting) & 1 \\
      \noalign{\smallskip}\hline\noalign{\smallskip}
      Information Leakage & None & None & None & Decontextualized K\&V & Forbidden Self-attention\\
      \noalign{\smallskip}\whline
   \end{tabular}}   
\end{table*}

\section{Gradient analysis}

\begin{figure}[htp]
   \begin{center}
      \includegraphics[width=0.45\textwidth]{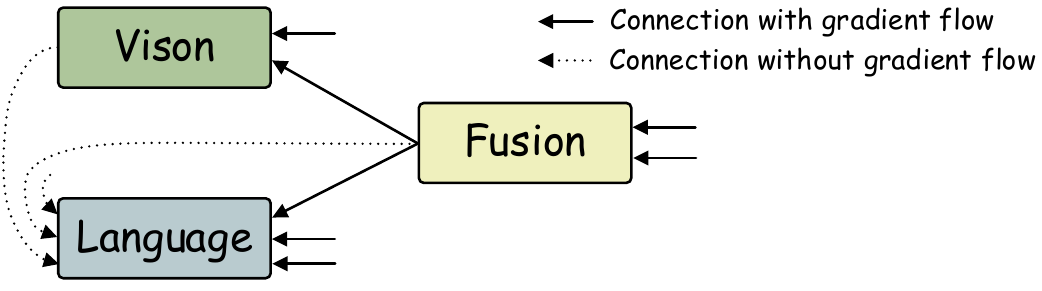}
      \caption{Analysis of gradient flow in ABINet++.}
      \label{fig:gradient}
   \end{center}
\end{figure}

As shown in Fig.~\ref{fig:gradient}, both the vision and language models receive gradient from either direct supervision or backpropagation of the fusion model, which ensures the independence of each model. Besides, the iteration can be regarded as a data augmentation method during the training phase, which has no disturbance on the flow of the gradient.

\section{Datasets}

\subsection{Scene Text Recognition}

\noindent\textbf{ICDAR 2013 (IC13)}~\cite{karatzas2013icdar} has 1,095 testing images. Among these images, 1,015 and 857 images are used for evaluation, where the 1,015 images are obtained by filtering non-alphanumeric characters, and the 857 images are further remained after pruning the texts shorter than 3 characters~\cite{wang2011end}.

\noindent\textbf{Street View Text (SVT)}~\cite{wang2011end} contains 647 testing images. Most of them are noisy, blurry, or low-resolution images. 

\noindent\textbf{IIIT 5K-Words (IIIT)}~\cite{mishra2012scene} includes 3,000 images for evaluation crawled from Google image search.

\noindent\textbf{ICDAR 2015 (IC15)}~\cite{karatzas2015icdar} contains 2,077 testing images. Following the previous works~\cite{baek2019wrong,cheng2017focusing}, a subset of the images in the number of 1,811 is also used for evaluation by discarding non-alphanumeric text images, and some vertical and extremely distorted images.

\noindent\textbf{Street View Text-Perspective (SVTP)}~\cite{quy2013recognizing} has 645 images for evaluation. Different from SVT, images in SVTP encounter severe perspective distortions.

\noindent\textbf{CUTE80 (CUTE)}~\cite{risnumawan2014robust} mainly contains curved text images. There are 288 word-box images for evaluation.

\noindent\textbf{MJSynth (MJ)} is a synthetic dataset~\cite{jaderberg2014synthetic,jaderberg2016reading}, which contains about 8.92M word-box images. All of the images regardless of the training, validation and testing splits are used as our training set.

\noindent\textbf{SynthText (ST)} is another commonly used synthetic dataset~\cite{gupta2016synthetic}, whose images are generated in full scene. About 6.98M word-box images are directly cropped from full scene images according to the ground-truth bounding boxes.

\noindent\textbf{Uber-Text}~\cite{Ying2017UberText} contains scene images collected from Bing Maps. We directly crop 571,534 word-box images, and construct an unlabeled dataset by removing all the annotations.

\subsection{Scene Text Spotting}

\subsubsection{Evaluation Datasets}

\noindent\textbf{Total-Text}~\cite{ch2020total} is a primary dataset for arbitrarily-shaped text spotting. This dataset consists of 1,255 training images and 300 test images. All the text instances are annotated by a 10-points polygon at the word-level.

\noindent\textbf{SCUT-CTW1500}~\cite{liu2019curved} is a widely-used dataset for arbitrarily-shaped text spotting. Compared with Total-Text, SCUT-CTW1500 is more challenging which contains long text with up to 100+ characters for a text instance. In addition, the text density is higher that small text may stack together in a large amount. There are 1,500 images in total, with 1,000 for training and 500 for evaluation. The annotations are labeled at line-level with a 14-points polygon for each text instance.

\noindent\textbf{ICDAR 2015}~\cite{karatzas2015icdar} is a popular multi-oriented dataset. ICDAR 2015 includes 1,000 training images and 500 testing images. Images are all annotated by word-level quadrilaterals.

\noindent\textbf{ICDAR 2019 ReCTS}~\cite{zhang2019icdar} is a challenging Chinese signboard dataset. There are 20K images for training and 5K images for evaluation. Images in this dataset are annotated at line-level and character-level by quadrilaterals.

\subsubsection{Training Datasets}

\noindent\textbf{SynthText150K}~\cite{liu2020abcnetv2,liu2020abcnet} is a synthetic dataset generated in the same way as SynthText~\cite{gupta2016synthetic}. Differently, SynthText150K contains more images, especially arbitrarily-shaped scene text images. There are 150K images in SynthText150K, and about 54K images contain curved text mostly.

\noindent\textbf{SynChinese130K}~\cite{liu2020abcnetv2} is released together with SynthText150K using the same synthetic engine. This dataset mainly focuses on Chinese text with about 130K images.

\noindent\textbf{ICDAR 2017 MLT}~\cite{nayef2017icdar2017} is a multi-lingual dataset, which is split into 7.2K training images, 1.8K validation images and 9K testing images. The images are annotated at word-level quadrangles.

\noindent\textbf{ICDAR 2019 ArT}~\cite{chng2019icdar2019} is an arbitrarily-shaped dataset containing English and Chinese text. ArT is composed of 5,603 training images and 4,563 testing images, which are labeled in tight polygons.

\noindent\textbf{ICDAR 2019 LSVT}~\cite{sun2019icdar} contains 450K street view images in total. Among them, 30K images are labeled as training set and 20K images are labeled as testing set. The labeled images are annotated in quadrilaterals with 8 or 12 vertexes.

\section{Analysis of Successful/Failure Cases for Scene Text Recognition}

\begin{figure*}
   \begin{center}
      \includegraphics[width=1.0\textwidth]{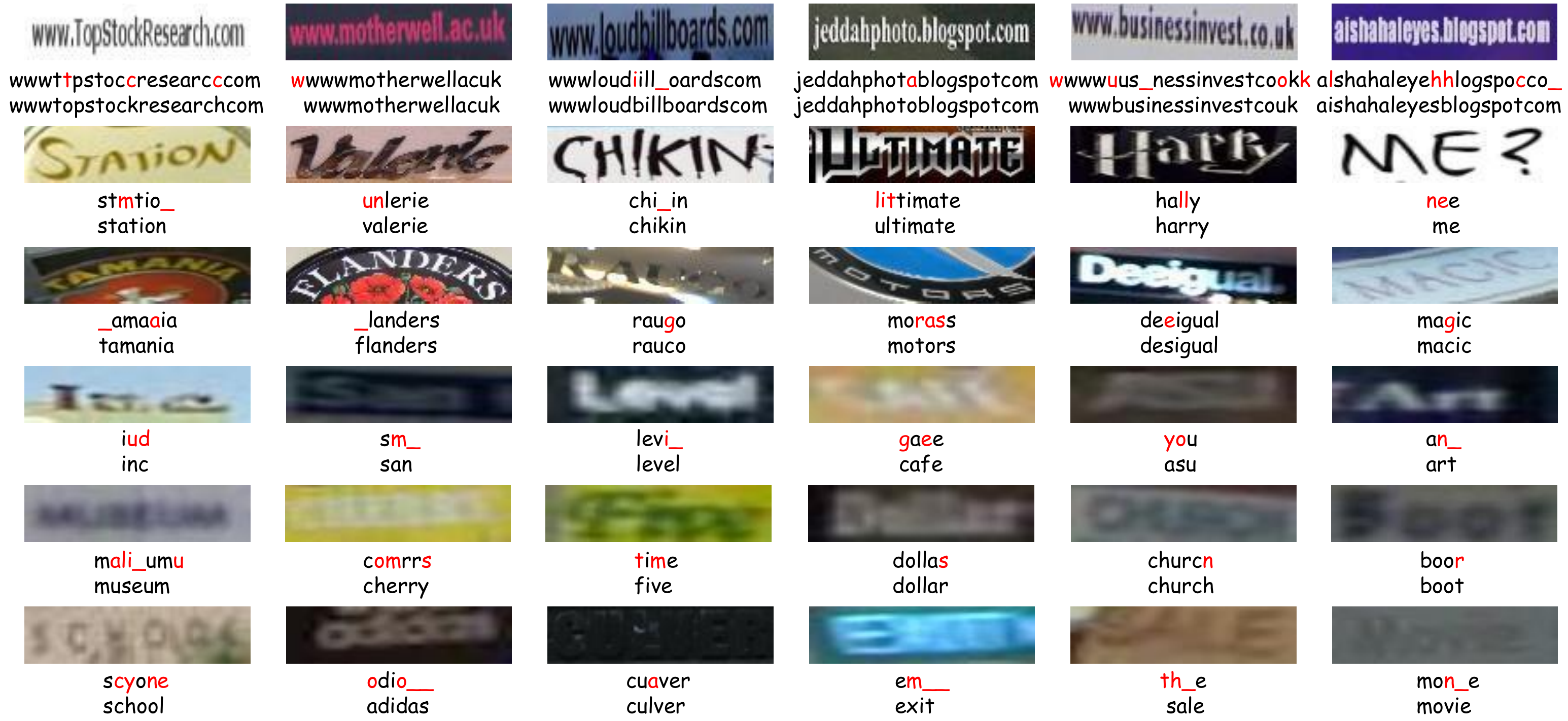}
      \caption{Successful and failure examples in scene text recognition.}
      \label{fig:semi_all_img}
   \end{center}
   \vspace{1em}
 \end{figure*}

 \begin{figure*}
   \begin{center}
      \includegraphics[width=1.0\textwidth]{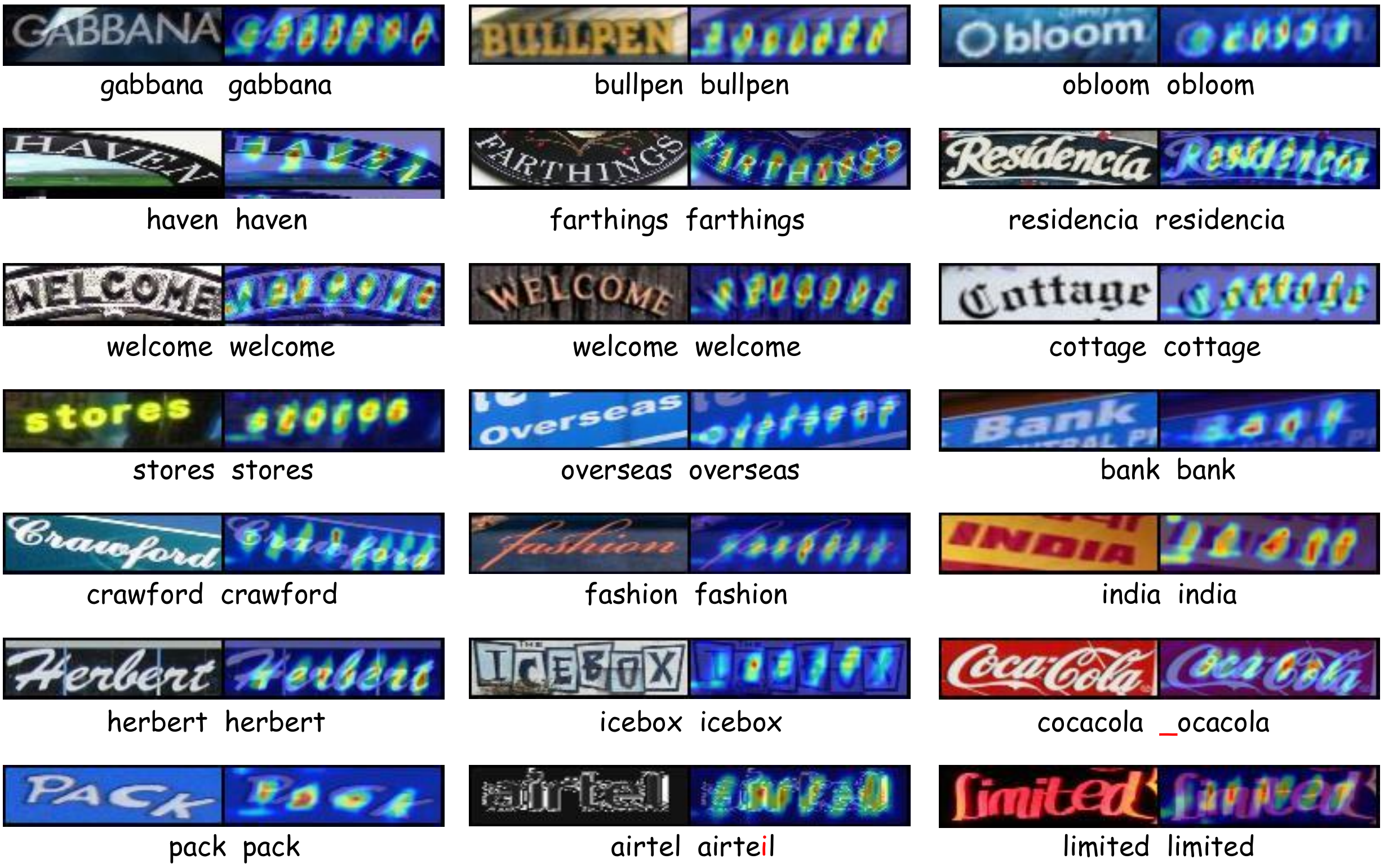}
      \caption{Visualization of attention maps for ABINet++ (LV).}
      \label{fig:attention_full}
   \end{center}
   \vspace{1em}
 \end{figure*}

Some examples successfully recognized by ABINet++$^\ddag$ (LV) but wrongly recognized by ABINet++ (LV) are given on the top line and bottom line respectively in Figure~\ref{fig:semi_all_img}. Failure cases of ABINet++ (LV) may come from extremely long text, unusual fonts, irregular text, blurred images, etc. We directly resize all the images to $32 \times 128$, which is unfriendly to extremely long text due to the squeezed visual patch. However, we observe that learning from more data using a semi-supervised way can alleviate this problem obviously. Besides, even though we find that the language model (BCN) can assist the recognition of the vision model in dealing with unusual fonts, learning from more text images with various fonts is still a straightforward and effective way. We also find that some irregular text that is not recognized by ABINet++ (LV) can be resolved by semi-supervised learning, which is the reason why ABINet++$^\ddag$ (LV) achieves significant performance on CUTE dataset (Table 7 in regular manuscript). In the last three rows of Figure~\ref{fig:semi_all_img}, we visualize some hard examples caused by blurred images, which indicates that ABINet++$^\ddag$ (LV) is able to recognize text under a hostile environment that even humans cannot read.

\section{Experiment on Chinese Dataset for Scene Text Recognition}

\begin{table}[htp]
   \begin{center}
   \caption{Recognition accuracy on ICDAR2015 TRW.}
   \label{tab:chinese}
   \begin{tabular}{|l|c|c|}
   \hline
   Method & Character Accuracy(\%)\\
   \hline
   CASIA-NLPR\cite{zhou2015icdar}& 72.1 \\
   SCCM w/o LM\cite{yin2017scene}& 76.5 \\
   SCCM\cite{yin2017scene}& 81.2 \\
   2D-Attention~\cite{yu2020towards}& 72.2 \\
   CTC~\cite{yu2020towards}& 73.8 \\
   SRN~\cite{yu2020towards}& 85.5 \\
   \hline
   ABINet++ & \textbf{87.1} \\
   \hline
   \end{tabular}
   \end{center}
 \end{table}

To validate the performance on non-Latin recognition, an additional experiment on ICDAR2015 Text Reading in the Wild Competition dataset (TRW15)~\cite{zhou2015icdar} is conducted. Experimental setup follows the configuration of SRN~\cite{yu2020towards}. Specifically, ABINet++ is trained on a synthetic dataset with 4 million images and the training sets of RCTW~\cite{shi2017icdar2017} and LSVT~\cite{sun2019chinese}. From the results in Table~\ref{tab:chinese} we can see, ABINet++ obtains a $1.6\%$ improvement on TRW15 dataset compared with SRN, showing that our ABINet++ is also robust to Chinese text recognition.

\section{Attention Maps}

Some examples of attention maps are given in Fig.~\ref{fig:attention_full}, where ground-truth text and the predictions of ABINet++ (LV) are listed. As can be seen from the visualization, the proposed method is able to accurately attend to the character regions.

\section{Analysis of Successful/Failure Cases for Scene Text Spotting}

\begin{figure*}[htp]
   \begin{center}
      \includegraphics[width=1.0\textwidth]{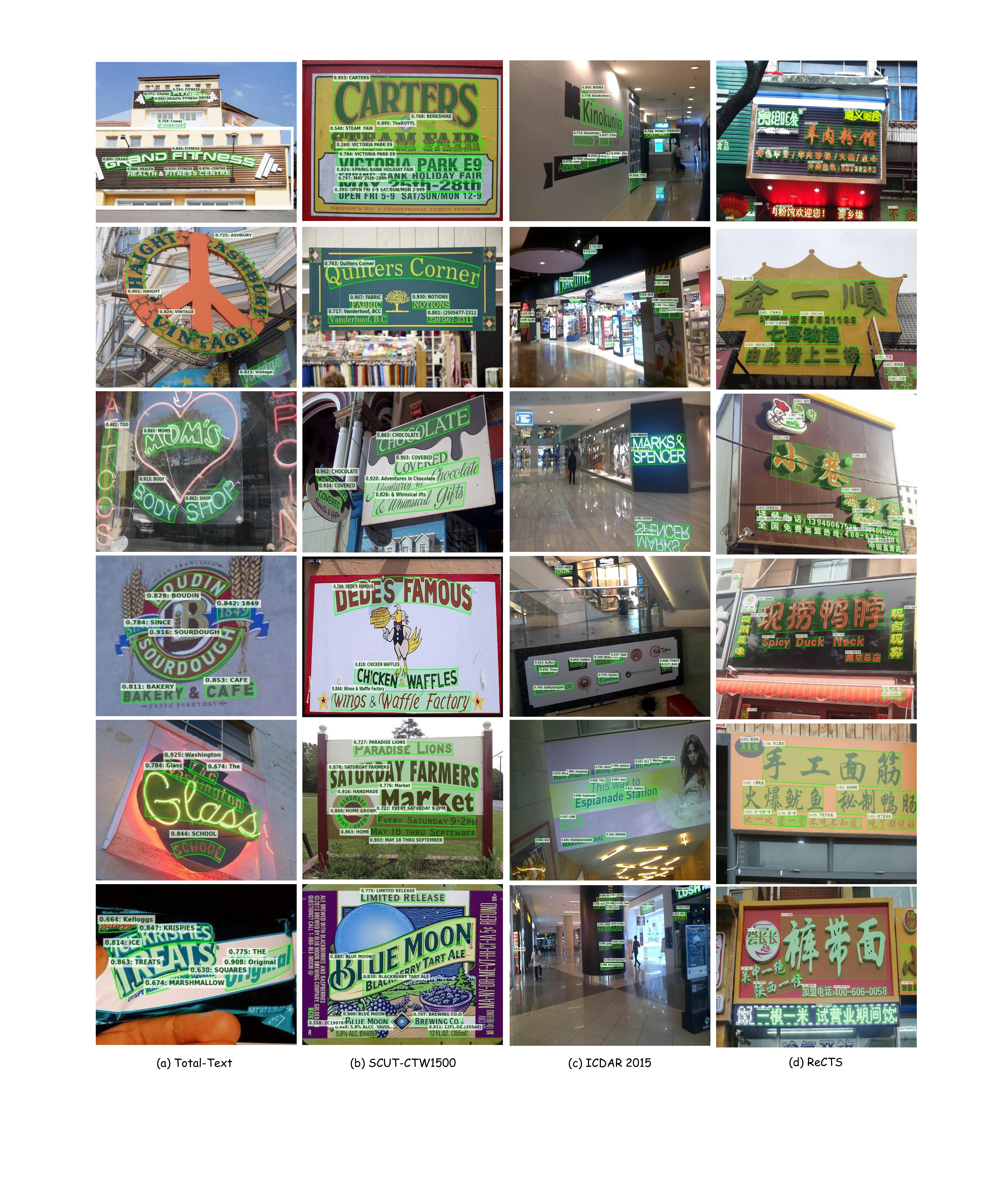}
      \caption{Successful and failure examples in scene text spotting.}
      \label{fig:e2e_img_full}
   \end{center}
   \vspace{-2em}
\end{figure*}

Fig.~\ref{fig:e2e_img_full} gives more examples on Total-Text, SCUT-CTW1500, ICDAR 2015 and ReCTS, including both successful and failure cases. 
 
{
\bibliographystyle{IEEEtran}
\bibliography{IEEEabrv_sup}
}